\documentclass{article}

\usepackage{PRIMEarxiv}

\usepackage[utf8]{inputenc} 
\usepackage[T1]{fontenc}    
\usepackage{hyperref}       
\usepackage{url}            
\usepackage{booktabs}       
\usepackage{amsfonts}       
\usepackage{nicefrac}       
\usepackage{microtype}      
\usepackage{lipsum}
\usepackage{fancyhdr}       
\usepackage{graphicx}       
\graphicspath{{media/}}     
\usepackage{subfigure}
\usepackage{algorithm}
\usepackage{algorithmic}
\usepackage{color}

\usepackage{colortbl} 
\usepackage{xcolor}
\usepackage{array}

\title{Uncertainty-Guided Depth Fusion for Spike Camera
}

\author{
  Jianing Li \\
  Nanjing University \\
  \texttt{jnli2021@gmail.com}
   \And
  Jiaming Liu \\
  Peking University \\
  \texttt{liujiaming@bupt.edu.cn} 
   \And
  Xiaobao Wei \\
  Beihang University \\
  \texttt{weixiaobao@buaa.edu.cn} 
    \And
  Jiyuan Zhang \\
  Peking University\\
  \texttt{jyzhang@stu.pku.edu.cn} 
    \And
  Ming Lu \\
  Intel Labs\\
  \texttt{lu199192@gmail.com} 
    \And
  Lei Ma \\
  Peking University\\
  \texttt{lei.ma@pku.edu.com} 
    \And
  Li Du \\
  Nanjing University\\
  \texttt{ldu@nju.edu.cn} 
    \And
  Tiejun Huang\\
  Peking University\\
  \texttt{tjhuang@pku.edu.cn} 
    \And
  Shanghang Zhang \\
  Peking University \\
  \texttt{shzhang.pku@gmail.com}\\
}

\begin{document}
\maketitle

\begin{abstract}
Depth estimation is essential for various important real-world applications such as autonomous driving. However, it suffers from severe performance degradation in high-velocity scenario since traditional cameras can only capture blurred images. To deal with this problem, the spike camera is designed to capture the pixel-wise luminance intensity at high frame rate. However, depth estimation with spike camera remains very challenging using traditional monocular or stereo depth estimation algorithms, which are based on the photometric consistency. In this paper, we propose a novel Uncertainty-Guided Depth Fusion (UGDF) framework to fuse the predictions of monocular and stereo depth estimation networks for spike camera. Our framework is motivated by the fact that stereo spike depth estimation achieves better results at close range while monocular spike depth estimation obtains better results at long range. Therefore, we introduce a dual-task depth estimation architecture with a joint training strategy and estimate the distributed uncertainty to fuse the monocular and stereo results. In order to demonstrate the advantage of spike depth estimation over traditional camera depth estimation, we contribute a spike-depth dataset named CitySpike20K, which contains 20K paired samples, for spike depth estimation. UGDF achieves state-of-the-art results on CitySpike20K, surpassing all monocular or stereo spike depth estimation baselines. We conduct extensive experiments to evaluate the effectiveness and generalization of our method on CitySpike20K. To the best of our knowledge, our framework is the first dual-task fusion framework for spike camera depth estimation. Code and dataset will be released.
\end{abstract}


\section{Introduction}
Depth estimation has shown great significance in many real-world applications, including robotic manipulation\cite{tremblay2018deep}, augmented reality \cite{tang20193d,marchand2015pose}, and autonomous driving \cite{manhardt2019roi,wu20196d}. However, it suffers from bottlenecks in high-velocity motion circumstances, hindered by blurred images from traditional low frame rate cameras \cite{hu2021optical}. To deal with high-velocity motion, spike cameras are designed to capture the images at high frame rate \cite{dong2019efficient,zhu2019retina}. Since spike cameras can capture the pixel-wise luminance intensity at high frame rate, spike depth estimation is an ideal solution to depth estimation in high-velocity motion \cite{zhu2020retina}.

\begin{figure}[t]
\begin{center}
	\includegraphics[width=8.5cm, height=6.5cm]{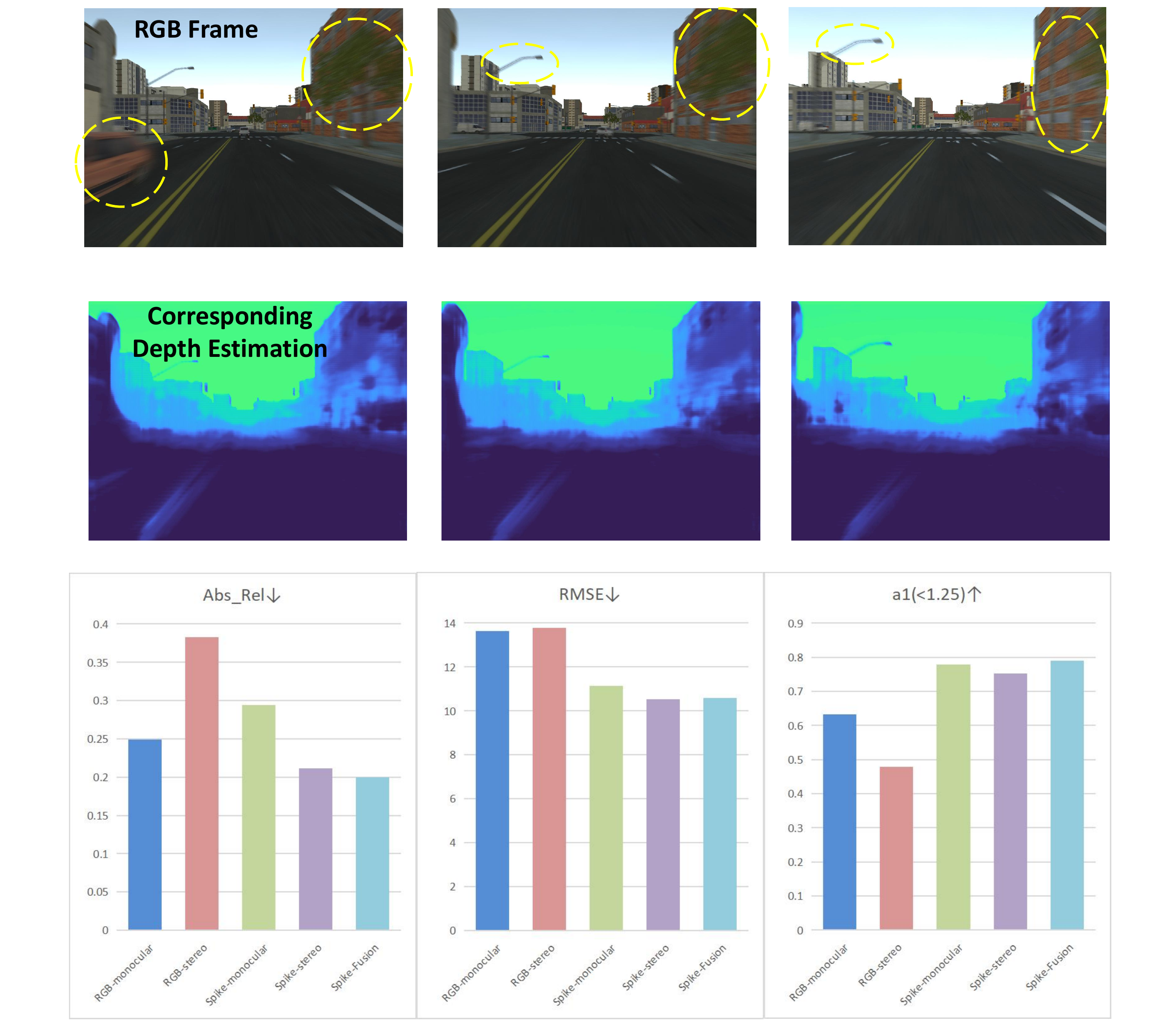}
\end{center}
\caption{We demonstrate the advantage of spike camera when dealing with fast-moving objects for driving depth estimation. The first row indicates the motion blur (yellow dotted circle) from traditional RGB camera. The second row indicates such motion blur brings inaccurate depth estimation for such high-speed objects.
The third row demonstrates the performance decrease for RGB depth estimation, compared with spike depth estimation. Therefore, we introduce spike camera and our proposed UGDF to solve this problem.}
\label{fig:0}
\end{figure}

\begin{figure}[t]
\begin{center}
	\includegraphics[width=8.5cm, height=4cm]{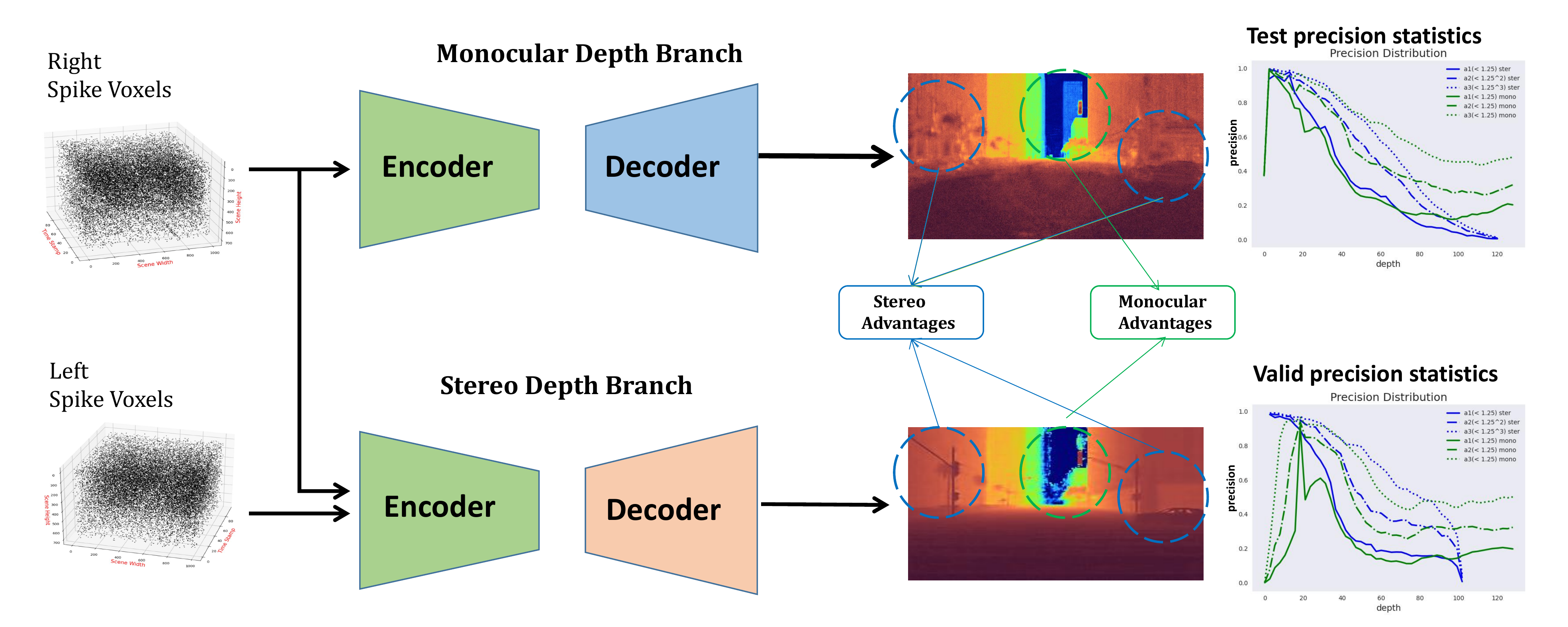}
\end{center}
\caption{ We use the binocular spike data as the input and train the monocular and stereo depth estimation models separately. By analyzing the predictions of monocular and stereo models, we find they have different accuracies at different depth ranges. This motivates us to fuse the predictions in a dual-task depth estimation architecture.}
\label{fig:0}
\end{figure}

Although there are plenty of traditional works on monocular depth estimation \cite{mayer2016large,kendall2017end,Khamis_2018_ECCV,Chabra_2019_CVPR,guo2019group} and stereo depth estimation \cite{Xu_2018_CVPR,Ramamonjisoa_2020_CVPR,lee2019monocular,Ramamonjisoa_2019_ICCV,fu2018deep,godard2017unsupervised}\cite{liu2022local}. It is still very challenging to apply them to spike depth estimation since spike data lacks reliable photometric consistency. In order to solve this problem, we first analyze the pros and cons of monocular and stereo depth estimation. On the one hand, monocular depth estimation is inherently ill-posed and mainly depends on the semantic knowledge of features. Therefore, it is robust to the disparity error and achieves better results at long range. On the other hand, stereo depth estimation compares the local patch pairs to obtain the optimal disparity. Therefore, it obtains better results at close range and performs worse at long range. As shown in Figure 1, we conduct the analysis of monocular and stereo spike depth estimation. This motivates us to fuse the monocular and stereo predictions for spike depth estimation, alleviating the problem of lacking reliable photometric consistency. 

In this paper, we propose a novel Uncertainty-Guided Depth Fusion (UGDF) framework to fuse the predictions of monocular and stereo spike depth estimation. Instead of training the monocular and stereo models separately, UGDF introduces a depth estimation architecture for dual tasks with a joint training strategy. This architecture includes two components. The first component is a shared encoder, which learns a feature representation to build stereo cost volume and monocular depth regression. The second component consists of two parallel branches for monocular and stereo depth estimation tasks. For the monocular branch, we set decoder to consist of three upsampling blocks. As for the stereo branch, we utilize a 3D hourglass-shaped convolution to aggregate the disparity dimension feature of 4D cost volume \cite{chang2018pyramid}. To fuse the predictions of both branches, instead of naive linear fusion, we introduce a novel adaptive uncertainty-guided fusion approach. Different from occlusion-aware fusion \cite{chen2021revealing}, which only exploits the knowledge from stereo branch, we adopt regression uncertainty formulations \cite{zhou2021sub} to measure the performances of monocular and stereo branches. Guided by the uncertainty maps, we fuse the reliable predictions of monocular and stereo branches, taking advantage of both tasks for the final estimation. 

In addition, we contribute a spike-depth dataset named CitySpike20K, which consists of 20K paired samples, for spike depth estimation. We demonstrate the great advantages of spike camera for high-velocity depth estimation on CitySpike20K. Extensive experiments are conducted to demonstrate the good performance of our framework compared with state-of-the-art monocular and stereo baselines.

Our contributions can be concluded as follows:
\begin{itemize}
\item We propose a novel Uncertainty-Guided Depth Fusion framework to fuse the predictions of monocular and stereo spike depth estimation, alleviating the problem of lacking reliable photometric consistency for spike data.

\item We introduce a dual-task depth estimation architecture along with a joint training strategy. To the best of our knowledge, we are the first to fuse dual tasks for spike depth estimation. 

\item We contribute a spike dataset named CitySpike20K, which contains 20K spike-depth pairs, to demonstrate the advantages of spike camera over traditional cameras on high-velocity depth estimation.

\item We conduct extensive experiments to evaluate the advantages of our method against existing monocular and stereo baselines.
\end{itemize}

\section{Related Work}

In this section, we investigate and reviewed recent works that are related to ours concerned with frame-based and event-based vision for depth estimation.
\subsection{Monocular and Stereo Depth Estimation}

Monocular and stereo methods are two parallel mainstreams in the development of depth estimation algorithms.
One of the earliest works that inspired recent trends for monocular depth estimation was introduced by Eigen et at.\cite{eigen2014depth}. This work proposed a kind of novel architecture that includes coarse-scale and fine-scale two steps, defining depth estimation as a pixel-wise regression problem. Similar to semantic segmentation task, one popular design for monocular depth estimation is encoder-decoder structure with  CNNs\cite{Xu_2018_CVPR,Ramamonjisoa_2020_CVPR,lee2019monocular,Ramamonjisoa_2019_ICCV,fu2018deep,godard2017unsupervised} or transformers\cite{ranftl2021vision,yang2021transformer}. In the encoding stage, the encoder captures context information and learns a global representation. And in the decoding stage, the network tends to establish a coupling connection between context texture and depth information with ground-truth full-supervision or self-supervision\cite{godard2017unsupervised,guizilini20203d,lyu2020hr}. Innovation has also been made in regression style\cite{fu2018deep,bhat2021adabins,roy2016monocular}, for more efficient representation of depth information. A recent study shows great potential in combining monocular depth estimation as auxiliary tasks for semantic segmentation\cite{Jiao_2018_ECCV,hoyer2021three}. 

Stereo depth estimation shows quite a different design strategy from monocular ways. Early works concentrate mainly on stereo matching for left and right stereo pairs\cite{4270415,7989227}. After deep learning was applied to this task, a stereo depth estimation pipeline contains three main steps: (1) feature extraction (2) cost aggregation, and (3) disparity/depth regression. Thanks to 3D convolution and the proposal of 3D cost-volume\cite{mayer2016large,kendall2017end}, the whole pipeline can be constructed end-to-end\cite{Khamis_2018_ECCV,Chabra_2019_CVPR}. PSMNet\cite{chang2018pyramid} concatenates left-fight features to cost volume and perform hourglass-shaped 3D convolution to make aggregation. GwcNet\cite{guo2019group} uses correlation formulation to divide cost volume into groups, which decreases computation while improving prediction results. In addition, self-supervised methods also gain competitive performance\cite{Zhou_2017_ICCV}. 

Recent study shows a new trend for rethinking the connection between monocular and stereo depth estimation\cite{Tosi_2019_CVPR,chen2021revealing}. Besides, left-right consistency has been the main cue for unsupervised depth estimation works of monocular task\cite{godard2017unsupervised}.

\subsection{Depth Estimation for Event-based Vision}
Compared to the standard frame-based cameras, biologically-inspired event-based sensors capture visual information with low latency and minimal redundancy. Dynamic Vision Sensors (DVS) is a  kind of representative event-based camera. Compared with frame-based camera, DVS is capable of capturing motional objects. Zhu et al.\cite{zhu2018multivehicle} provides a 100Hz DVS dataset containing depth ground-truth. They also propose time-synchronized event disparity volumes in \cite{zhu2018realtime} to handle DVS data for stereo matching. Similar research\cite{zhu2019unsupervised} uses discretized event volume to supervise monocular optical flow and depth estimation without labels. Another work\cite{ranccon2021stereospike} adopts a spiking neural network to estimate event depth.

\subsection{Spike Camera and Its Visual Application  }
Spike camera is also a kind of novel bio-inspired event camera. Distinct from the frame-based cameras and dynamic vision sensors, spike camera mimics the retina to record the nature scenes by continuous-time spikes\cite{yu2020toward,zhu2019retina}. \cite{9181055} develops a new image reconstruction approach for potential retina-inspired spike camera to recover high-speed motion scenes. \cite{Zheng_2021_CVPR,Zhao_2021_CVPR,zheng2021high,zhu2021neuspike}use spiking or regular convolutional neural networks to reconstruct high quality and high-speed images like a human vision from spike streams. Spike vision shows obvious advantages in capturing high-speed moving objects or scenes, so it provides new solutions for some long-standing problems in the field of machine vision.


\begin{figure*}[t]
\includegraphics[width=17.2cm, height=7.2cm]{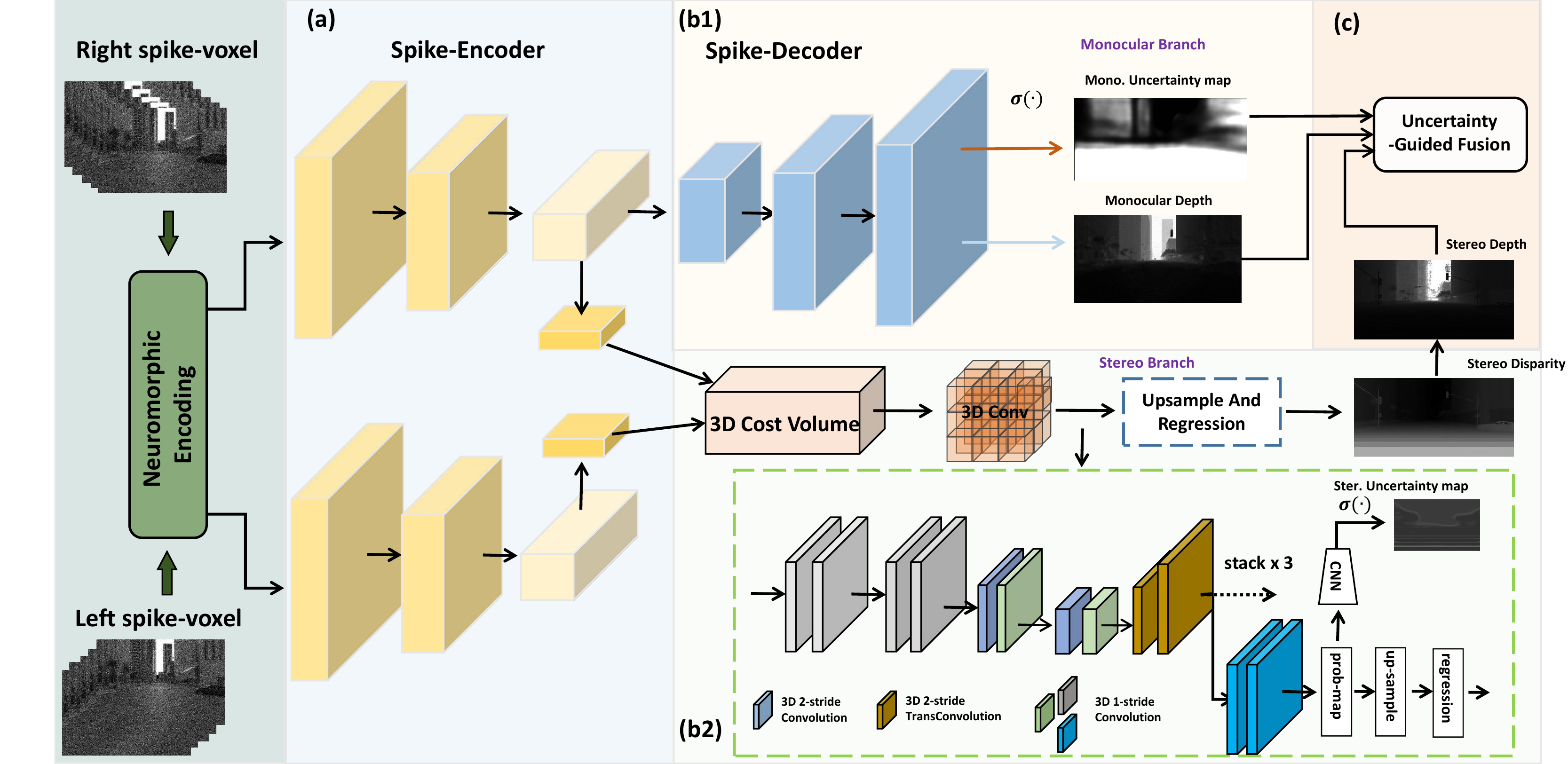}
\centering
\caption{\textbf{Illustration of the network architecture}. The network consists of three major modules. Processed spike data pairs are sent into spike encoders, which contain 3 downsampling layers, for initial representation (a). Monocular and stereo branches deal with these features, and output depth and disparity respectively (b1, b2). A final uncertainty-guided fusion is performed to aggregate monocular and stereo results (c).}
\label{fig:3}
\end{figure*}


\subsection{Spike Camera}
Different from the RGB cameras and dynamic vision sensors, spike camera mimics the retina to record natural scenes by continuous-time spikes \cite{yu2020toward,zhu2019retina}. \cite{9181055} develop a new image reconstruction approach for the spike camera to recover high-speed motion scenes. \cite{Zheng_2021_CVPR,Zhao_2021_CVPR,zheng2021high,zhu2021neuspike} use spike or regular CNNs to reconstruct high quality and high-speed images from spike streams. Spike vision shows obvious advantages in capturing high-speed moving objects or scenes, so it provides new solutions to some long-standing problems in the field of computer vision. In this paper, we propose a novel method for high-quality spike depth estimation by fusing monocular and stereo depth estimation.

\section{Proposed Method}

In this section, we present our method uncertainty-guided depth fusion framework (UGDF) to fully complement the strengths of both stereo and monocular tasks in spike data. The whole framework is demonstrated in Fig. \ref{fig:3} which consists of four components.


\subsection{Spike Data Analysis}
\label{sec3.1}
For spike camera, natural lights are captured by photoreceptors and converted to voltage under the integration of time series $t$. Once the voltage at a certain sensing unit reaches a threshold $\Theta$, a one-bit spike is fired and the voltage is reset to zero at the same time~\cite{9181055}.
\begin{equation}
   S(i, j, t) = \left\{
    \begin{array}{l}
            1 , \quad \int_{t0_{i,j}^{pre}}^{t} I(i, j) \, dt \ge \Theta\\  
            0 , \quad \int_{t0_{i,j}^{pre}}^{t} I(i, j) \, dt < \Theta
        \end{array}
\right.
\end{equation}
The above formula reveals the basic working pipeline of the spike camera, where $I(i,j)$ represents the luminance of pixel $(i,j)$, and $t0^{pre}_{i,j}$ represents the time that fires the last spike at pixel$(i,j)$. An ideal Analog to Digital Conversion(ADC) process is by continuous time, but such circumstances do not exist due to the inherent limitations of the digital circus. Even so, the spike camera is still able to generate much more dense frames than RGB models like streams, at a maximum frequency of 40000Hz\cite{Zhao_2021_ICCV,zhu2020retina, Zhao_2021_CVPR}.  Suppose we have $H \times W$ receptive field, the camera would output a $H \times W$ binary spike frame at a certain moment, and as time goes on, high-frequency spike frames are produced. However, directly performing depth estimation on spike frames remains challenging. On one hand, high contrast between 1-bit spike data makes it more difficult to distinguish local context information. On the other hand, different light intensities in the scene cause different frequencies of spike generation. So in practice, we make spike data in a fixed size time window to be a multi-channel tensor. For example, we take spike frames from continuous 100 time-steps frames and concatenate them at time dimension as $100 \times H \times W$ voxels, which then become inputs to our designed networks.

\subsection{UGDF Framework}
\label{sec3.2}
We propose a simple yet efficient network that includes a shared spike-encoder and a spike-decoder with two branches.
First, we build a neuromorphic encoding module to extract spiking features in both time domain and frequency domain. The spike voxel $V \in \mathbb{Z}^{100\times H\times W}$ is split into spike sequences $V_s = \{ v_1, v_2, ..., v_s \} $ by a fixed length of time window n, where s = $|100/n|_{\mathbb{Z}}$ and $v_s\in \mathbb{Z}^{n\times H\times W}$ . Then, the spike sequences are fed into a ConvRNN to extract temporal connections. Meanwhile, an FFT operation is performed on spike voxel $V$ to extract global information in frequency domain.

we use a shared deep encoder to learn a representation to build stereo cost volume and monocular depth regression, as shown in part (a) of Figure 3. We adopt MobileNetV3\cite{howard2019searching} as our encoder to make a trade-off between computation cost and model performance. The encoder contains 3 downsampling stage and the final feature map size of coding is $B \times 256 \times \frac{H}{8} \times \frac{W}{8}$.


As shown in the part (b1) and (b2) of Figure 3, two parallel branches stretch away for monocular and stereo depth estimation and serves as the spike-decoder. Different from any other previous works, we fuse monocular and depth estimation in one workflow at a multi-task level. In the stereo branch, we found a common ground for monocular and stereo tasks to learn a global context representation. So we take advantage of the encoder module, and concatenate the unary features(obtained coding from spike encoders) to build a 4D cost volume($ 256 \times \textit{Max-Disp.} \times \frac{H}{8} \times \frac{W}{8}$) for stereo disparity regression, where $\textit{Max-Disp.}$ represents the maximum disparity level to regression. Then inspired by \cite{chang2018pyramid}, we perform a 3D hourglass-shaped convolution to aggregate the disparity dimension feature of 4D cost volume. We stack three 3D hourglasses, each of which contains two blocks with \textit{$3\times3\times3$} kernel size and 2-stride 3D convolution, and two blocks with \textit{$3\times3\times3$} kernel size and 2-stride 3D transposed convolution. The disparity map is regressed at the final 3D convolution stage via a \textit{soft-argmin} operation\cite{kendall2017end}:
\begin{equation}
    \textit{soft-argmin:}=\sum_{d=0}^{Disp^{*}_{max}}{d \times \gamma{(-c_d)}}
\end{equation}
where $\gamma{(\cdot)}$ represents soft-max operation at disparity dimension, $c_d$ is predicted costs for disparity $d$, and $Disp^{*}_{max}$ means length of disparity dimension of the output features. The final disparity is weighed by a normalized probability. In the training phase, three 3D hourglass disparity outputs are all involved in building loss function. And the last output of three 3D hourglasses is used for the evaluation process. 

In the monocular branch, the right unary features(coding) are then sent to a decoding block for depth estimation. The decoder consists of three upsample blocks, each block contains one bilinear interpolation module and two convolutional layers along with batch normalization and Mish activation. The output layer is a $1\times1$ convolution which squeeze the features to \textbf{two channels}, \textbf{one} of which is used for \textbf{depth} estimation and \textbf{the other} one used for \textbf{uncertainty} allocation, as described in the next subsection.


\subsection{Uncertainty Guided Fusion}
\label{sec3.3}
Inspired by SUB-Depth\cite{zhou2021sub}, we assume the distribution over the output of either branch can be modeled as exponential family distribution such as Laplace's distribution or Gaussian distribution. The stereo branch and monocular branch adopt the same approach while we use the monocular branch as an illustration example. Given a dataset with left and right spike frames and corresponding depth ground truth $(x_l, x_r, y_l, y_r)$, we let our monocular branch output the mean $\hat{y}$ and variance $\sigma$ of the posterior probability distribution $p(y_r|\hat{y_r}, x_r)$. We use Laplace's distribution as:
\begin{equation}
    p(y_r|\hat{y_r}, x_r) = \frac{1}{2\sigma}exp\frac{-|\hat{y_r}-y_r|}{\sigma}
\end{equation}
We can convert the above distribution to a log-likelihood formula like:
\begin{equation}
    log(p(y_r|\hat{y_r}, x_r)) = -log(\sigma) + \frac{-|\hat{y_r}-y_r|}{\sigma} + const.
\end{equation}
So according to max-posterior probability estimation, an uncertainty loss can be formulated in the form of:
\begin{equation}
    loss_{unc.} = log(\sigma) + \frac{|\hat{y_r}-y_r|}{\sigma}
\label{eq:unc}
\end{equation}

We can minimize this loss function to obtain a max-posterior probability distribution over estimated monocular depth $\hat{y}$. The uncertainty coefficient $\sigma_m$ and predicted depth $\hat{y}$, which are the outputs of the part (b1), are regressed from the same decoder at the same time, so $\sigma_m$ can be seen as prediction uncertainty for monocular depth estimation task. Similar to the monocular branch, a lite CNN is added behind the probability map of the stereo branch, and regresses uncertainty coefficient $\sigma_s$ after a sigmoid activation.

We notice that the monocular branch outperforms the stereo branch in farther regions and the stereo branch is good at predicting closer regions. So we hope the fusion style may combine both monocular and stereo advantages. So we define a distance threshold:
\begin{equation}
    \sigma_{dis.} = D_{max} \frac{e^{2(\sigma_m - \sigma_s)} }{1 + e^{2(\sigma_m - \sigma_s)}} 
\label{eq:unc}
\end{equation}

\begin{figure}[t!]
\includegraphics[width=0.45\textwidth]{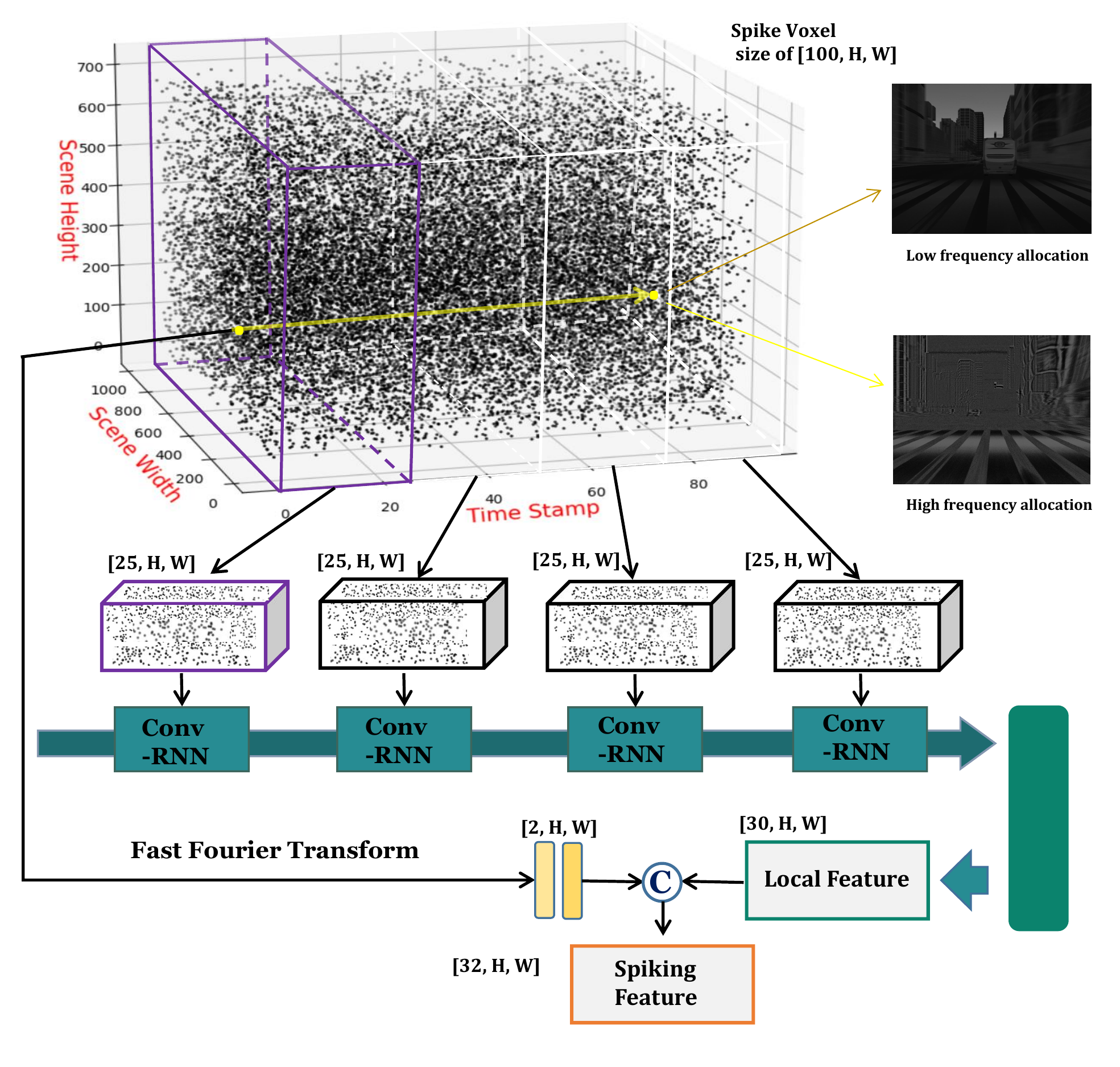}
\centering
\caption{Neuromorphic encoding module. We use Conv-RNN to extract local information, and meanwhile a Fast-Fourier-Transform is applied through the whole spike voxel to extract global information.}
\label{fig:5}
\end{figure}

With estimated uncertainty, an uncertainty-guided fusion mask $F$ can be defined as:

\begin{equation}
   F_i = \left\{
    \begin{array}{l}
            0 , \hat{D}_{mono.} \leq \sigma_{dis.}\\  
            1 , \hat{D}_{mono.} > \sigma_{dis.}
        \end{array}
\right.
\end{equation}

where i represents i-th element of the uncertainty map, and $\sigma_{dis}$ represents an uncertainty threshold to make fusion. 
To take advantage of both monocular and stereo branches, we use monocular depth prediction results $\hat{D}_{mono.}$ and stereo depth prediction results $\hat{D}_{ster.}$ to make further fusion, exploiting complementary advantages for monocular and stereo models. Instead of directly performing linear addition between two kinds of outputs, we fuse them in a more efficient uncertainty-guided way. And the uncertain-guided fusion is given:
\begin{equation}
    \hat{D_f} = F \odot \hat{D}_{mono.} + (1 - F) \odot \hat{D}_{ster.}
\end{equation}

\subsection{UGDF Loss Functions}
\label{sec3.4}
We present training strategies for baseline network without fusion and UGDF with fusion. The training loss of baseline network consists of monocular depth estimation \textit{$loss_{disp.}$} and stereo disparity regression \textit{$loss_{depth}$}, which use smooth-L1 loss during the training phase under the supervision of depth ground-truth and generated disparity labels. The baseline $loss_{base}$ is shown as below:\\

\begin{equation}
    loss_{base.} = loss_{disp.} + loss_{depth.}
\end{equation}
in which \textit{$loss_{disp.}$} and \textit{$loss_{depth.}$} are shown as below:
\begin{equation}
    loss_{disp.}(d^{*}, \hat{d}) = \frac{1}{N}\sum_{i=1}^{M}\sum_{j=1}^{N} \alpha_i \cdot smoothL1(d^{*}, \hat{d})
\end{equation}
\begin{equation}
    loss_{depth.}(D, \hat{D}) = \frac{1}{n}\sum_i{c_i}^2 - \frac{1}{n^2}(\sum_i{c_i})^2 + \eta
\end{equation}

In which $\eta = 0.1$ and $c_i = log(D_i) - log(\hat{D_i})$. And N is the size of data and M=3 is the number of stacked 3D hourglasses, $\left\{ \alpha_1, \alpha_2, \alpha_3  \right\}$ = $\left\{ 0.5, 0.7, 1.0  \right\}$. $\hat{D}$ and $D$ represent predicted depth and depth ground truth respectively. Similarly, $\hat{d}$ means predicted disparity and $d^{*}$ is generated disparity ground truth.

The training loss of UGDF consists of five losses, including monocular depth estimation \textit{$loss_{disp.}$}, stereo disparity regression \textit{$loss_{depth.}$}, monocular branch uncertainty \textit{$loss_{mono\_unc.}$}, stereo branch uncertainty \textit{$loss_{ster\_unc.}$} and fusion \textit{$loss_{fu.}$}. The whole UGDF \textit{$loss_{ugdf.}$} is shown as:
\begin{equation}
    loss_{ugdf.} = loss_{base.} + loss_{ster\_unc.} + loss_{mono\_unc.} 
\label{eq:ugde}
\end{equation}


where \textit{$loss_{mono-unc.}$} and \textit{$loss_{ster-unc.}$} follow Eq. \ref{eq:unc} while $\hat{D_f}$ denotes the fusion predicted depth from two branches. To be mentioned, the depth of the stereo branch is converted from disparity under intrinsic parameters of the camera. The training details of baseline and UGDF are the same. Further training details are presented in Section 5.

\section{Spike-Depth Dataset: CitySpike20K}
This section introduces different aspects of the dataset we propose. The dataset includes RGB scenes, and their corresponding spike frames and depth maps. All these data are generated by a simulated spike camera in Unity3D virtual environment. The dataset describes 11 sequences of city street scenes containing 6 day scenes and 5 night scenes.

Our proposed CitySpike20K dataset provides a depth estimation benchmark for spike data. The scenes are created via a simulated spike camera, recording a fast-moving car in the street scene, at a frequency of 1000Hz. The resolution for recorded data is $1024 \times 768$. We also build a depth field for these scenes and store them as 24-bit depth maps. The ground truth information is of 0.3-1000m absolute depth. In addition, we provide the focus $f$ and baseline length $base_{len}$ of the stereo camera in supplement. We also convert depth $D$ to disparity $disp$ under the function $disp. = \frac{f*base_{len}}{D}$. A visualization of our dataset is shown in \textbf{supplementary file}. Besides, in terms of sensor-collaboration, we provide 842 pairs of RGB images from regular stereo cameras, dense spike frames from stereo Vidar, as well as depth maps from stereo depth cameras. Three kinds of data are organized in a one-to-one corresponding way. Besides, we provide a demo sequence of 40000Hz frequency spike data, recording a 91km/h car driving in the city street. This demo is for  evaluating the depth estimation algorithm when loaded with high-frequency spike data.

\section{Experiments}

In this section, we conduct extensive experiments to show the advantages of UGDF. Then, we extensively evaluate UGDF by comparing it with the state-of-the-art and classic depth estimation methods which have shown great performance on RGB depth datasets such as KITTI\cite{Uhrig2017THREEDV} and NYUD-V2\cite{Silberman:ECCV12}. We also conduct comprehensive ablation studies to evaluate the contribution of each component in the last subsection. Due to space limitations, some details of experiments and results are provided in the supplementary materials.

\begin{table*}[t]
  \centering
  \scriptsize
  \caption{Quantitative results on CitySpike20K (decribed as CS20K below) \textbf{validation} set. Evaluation metrics are as described in section 3. We make comparison with GwcNet\cite{guo2019group}, CFNet\cite{shen2021cfnet}, PSMNet\cite{chang2018pyramid}. The evaluation metrics are as introduced in subsection 4.2. We also consider model parameter size to be one of the compared targets.} 
    \begin{tabular}{c|c|c|c|ccccccc}
    
    \toprule
    Dataset & Method & Approach & Modality &  \multicolumn{1}{l}{Abs\_Rel↓}\cellcolor{lightgray} & \multicolumn{1}{l}{RMSE ↓} & \multicolumn{1}{l}{Sq\_Rel ↓} & \multicolumn{1}{l}{RMSE\_log ↓} & \multicolumn{1}{l}{a1 ↑} & \multicolumn{1}{l}{a2 ↑} & \multicolumn{1}{l}{a3 ↑} \\
    \midrule

    & PSMNet & Ster.& RGB & 0.4564 & 15.484 & 12.990 & 0.734 & 0.469 & 0.668 &0.743 \\
  
    CS20K    & GwcNet & Ster. & RGB & 0.419 & 19.724 & 9.753 & 0.632 & 0.469 & 0.685& 0.767 \\
    
    & CFnet & Ster. & RGB & 0.4038 & 14.928 & 8.870 & 0.437 & 0.593  & 0.677&0.786 \\

    \midrule
   CS20k & \textbf{UGDF}(Ours) & Fusion & Spike & \textbf{0.2282 } &   \textbf{11.075 }   &   \textbf{4.699} &   \textbf{0.305}   &   \textbf{0.754}    & \textbf{0.879}&\textbf{0.942} \\
    \bottomrule
    \end{tabular}%
  \label{tab:1}%
\end{table*}%

\begin{table*}[t]
  \centering
  \scriptsize
  \caption{Quantitative results on CitySpike20K \textbf{test} set. We add two monocular algorithms as baselines which are DPT\cite{ranftl2021vision} and UNet\cite{ronneberger2015u}. } 
    \begin{tabular}{c|c|c|c|ccccccc}
    \toprule
    Dataset & Method & Approach & Modality &  \multicolumn{1}{l}{Abs\_Rel↓} & \multicolumn{1}{l}{RMSE ↓} \cellcolor{lightgray} & \multicolumn{1}{l}{Sq\_Rel ↓} & \multicolumn{1}{l}{RMSE\_log ↓} & \multicolumn{1}{l}{a1 ↑} & \multicolumn{1}{l}{a2 ↑} & \multicolumn{1}{l}{a3 ↑} \\
    \midrule

    & UNet & Mono.& RGB & 0.3612 & 19.217 & 6.981 & 0.502 & 0.569 & 0.765 & 0.893 \\

    & DPT & Mono.& RGB & 0.249& 13.641 & 4.349 & 0.379 & 0.632 & 0.817 & 0.925 \\
    
   CS20K & PSMNet & Ster.& RGB & 0.4341& 16.294 & 9.247 & 0.840 & 0.411 & 0.626 & 0.712 \\

     & GwcNet & Ster. & RGB & 0.3931 & 18.680 & 8.745 & 0.577 & 0.492 & 0.704& 0.787 \\

    & CFnet & Ster. & RGB & 0.3825 & 13.794 & 7.925 & 0.496 & 0.467  & 0.723& 0.836 \\

    \midrule
   CS20k & \textbf{UGDF}(Ours) & Fusion & Spike & \textbf{0.1997 } &   \textbf{10.953 }   &   \underline{4.879} &   \underline{0.412}   &   \textbf{0.790}    & \textbf{0.888}&\textbf{0.945} \\
    \bottomrule
    \end{tabular}%
  \label{tab:1}%
\end{table*}%

\subsection{Implementation Details}
\label{sec:5.1}

We train our proposed UGDF network on spike-depth pairs, including stereo spike frames and right depth-ground-truth.
The whole training phase contains 200 epochs and takes about 16 hours with the batch size of 4 on two NVIDIA-Tesla P100 GPUs, for $256 \times 512$ resolution spiking frames.

We utilize 24-bit 0-1000m absolute depth ground-truth to supervise training for the monocular branch. We normalize depth ground truth $D$ to $D^{*} \in (0,1)$, with the function $ D^{*} = D / 1000 $. Meanwhile, the disparity is transformed from depth with camera intrinsics.

As for optimization, we use Adam optimizer with $(\beta_1, \beta_2) = (0.9, 0.999)$. We set an initial learning rate of 1e-3 and decay to 0.33e-3 at epoch 35 for the sake of a more smooth optimizing process.

In this section, we compare UGDF against the state-of-the-art and classic depth estimation methods on CitySpike20K dataset.

\textbf{Data processing} Our proposed dataset contains 20K frames of spike-depth pairs. We split 7 out of 10 total sequences for training, 2 sequences for testing and 1 sequence for validating. All the data used for training and validating is sampled every 100 time-stamps to form a spike voxel. So we obtain 140 training pieces and 40, 20 for testing and validating our framework. To emphasize the advantage of spike data, we use blurred 30fps RGB frames in our dataset to train the RGB-based baseline methods. So there are 571 training pieces, 142 testing pieces and 111 validation pieces for baseline methods.

\textbf{Baseline methods} To demonstrate the effectiveness of UGDF, we compare it with some state-of-the-art and classic depth estimation methods which have shown remarkable performance on our proposed CitySpike20k dataset. For monocular methods, we choose classical UNet \cite{ronneberger2015u} and DPT\cite{ranftl2021vision}. UNet has been demonstrated a successful design on semantic segmentation\cite{ronneberger2015u, baheti2020eff} and image reconstruction\cite{Chen_2021_CVPR, chen2021hinet}. we adopt its proposed structure and evaluate it on our spike-depth estimation task. In DPT, we use Vit-b16 as the backbone and 224x224 as input resolution. PSMNet\cite{chang2018pyramid} uses subtract and concatenation method to build a 3D cost volume. GwcNet\cite{guo2019group} proposes group-wise correlation to reduce computation while conducting 3D convolution. CFNet\cite{shen2021cfnet} employs a variance-based uncertainty estimation to adaptively search disparity space. 
    
\begin{figure*}[t]
  \centering
    \subfigure[RGB Frame]{\includegraphics[width=0.15\textwidth]{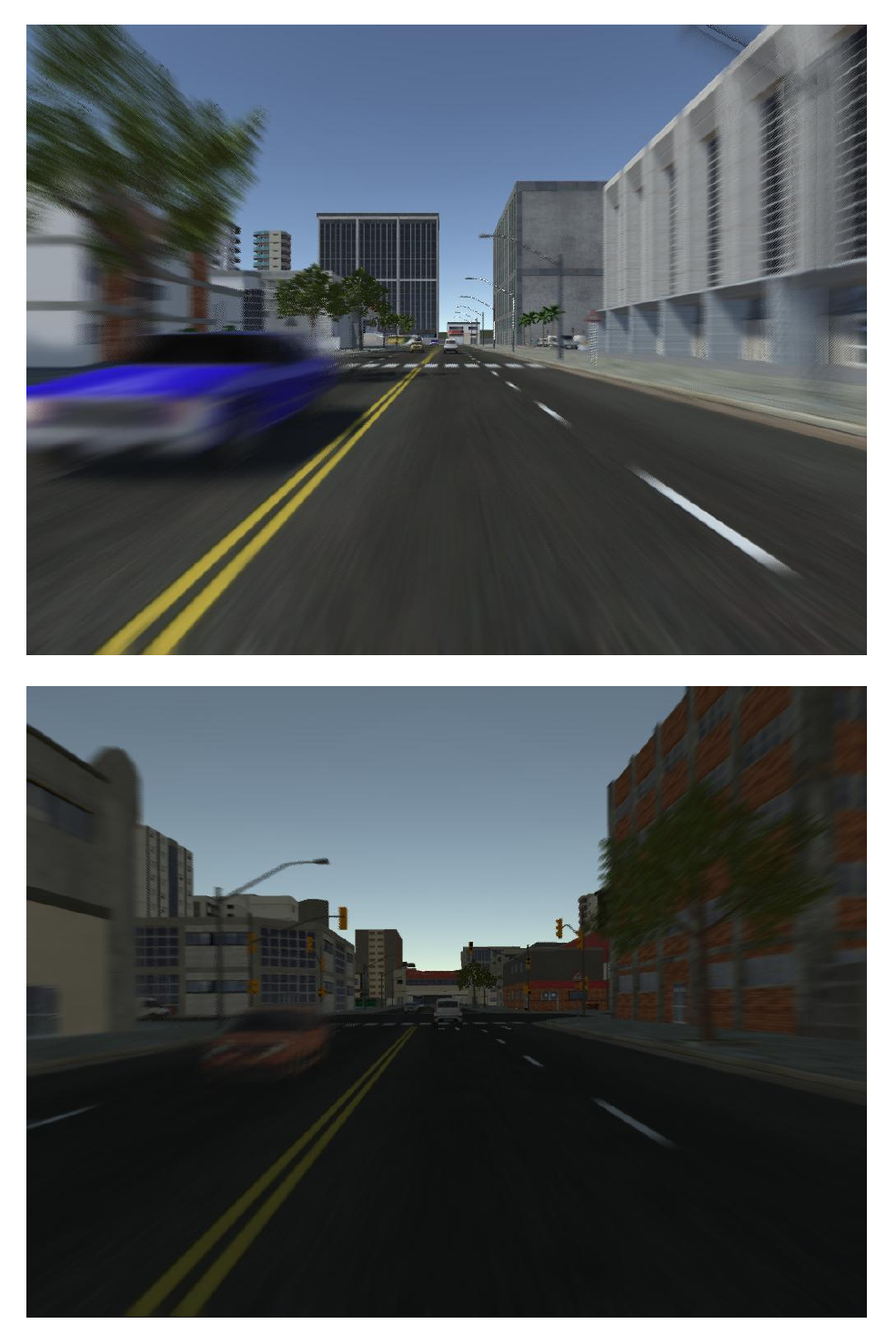}} 
	\subfigure[Spike Frame]{\includegraphics[width=0.15\textwidth]{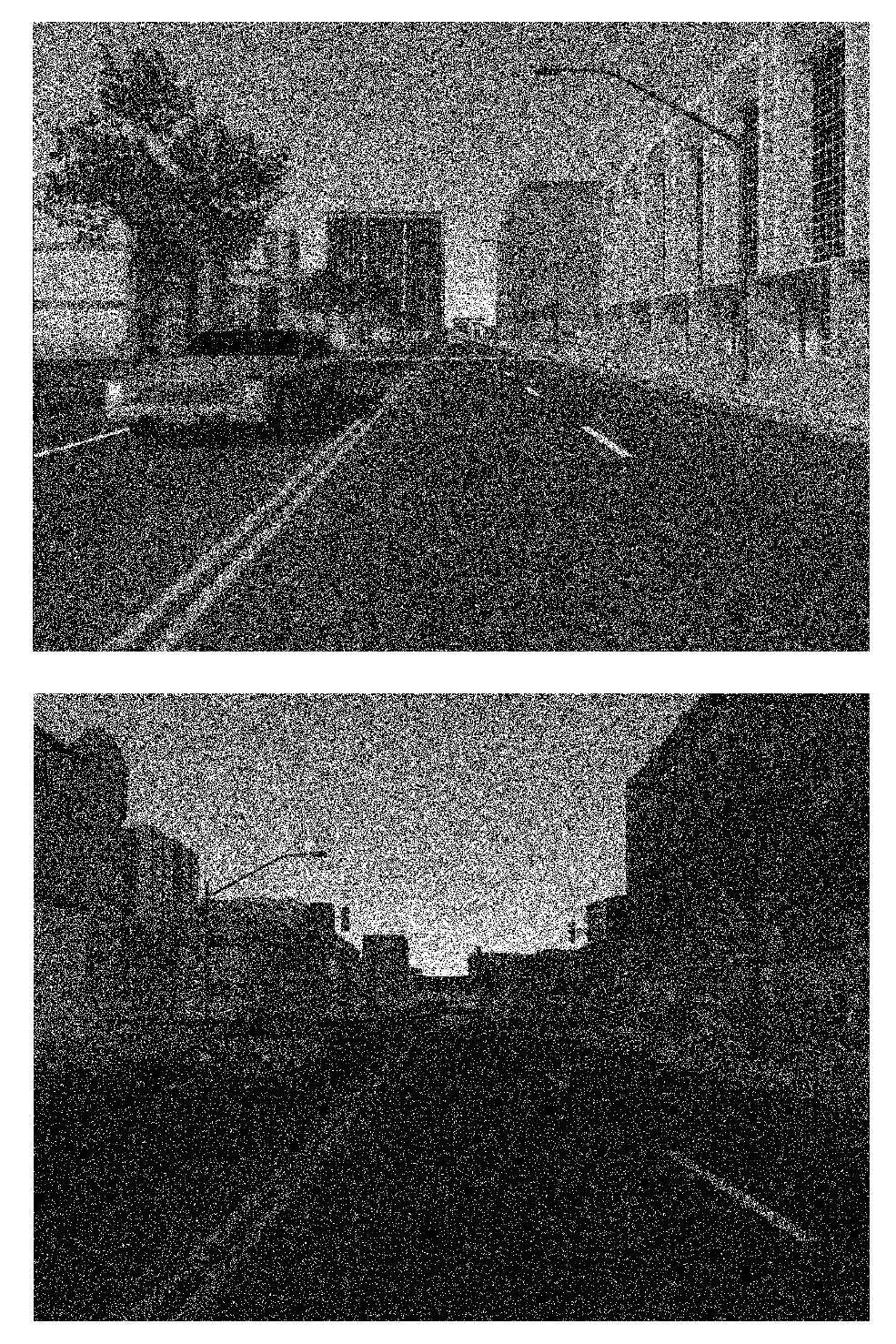}} 
    \subfigure[Mono. result]{\includegraphics[width=0.15\textwidth]{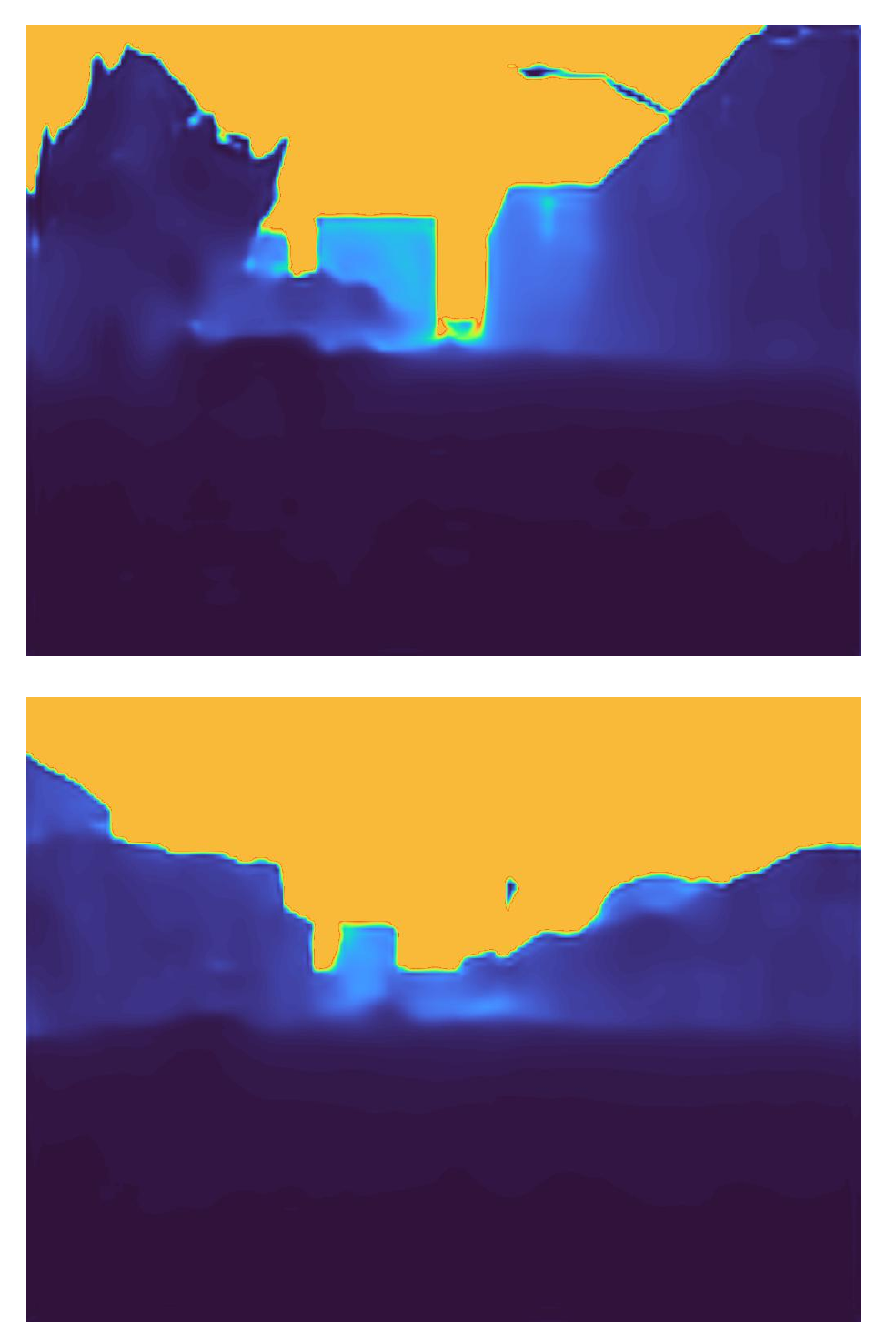}} 
    \subfigure[Ster. result]{\includegraphics[width=0.15\textwidth]{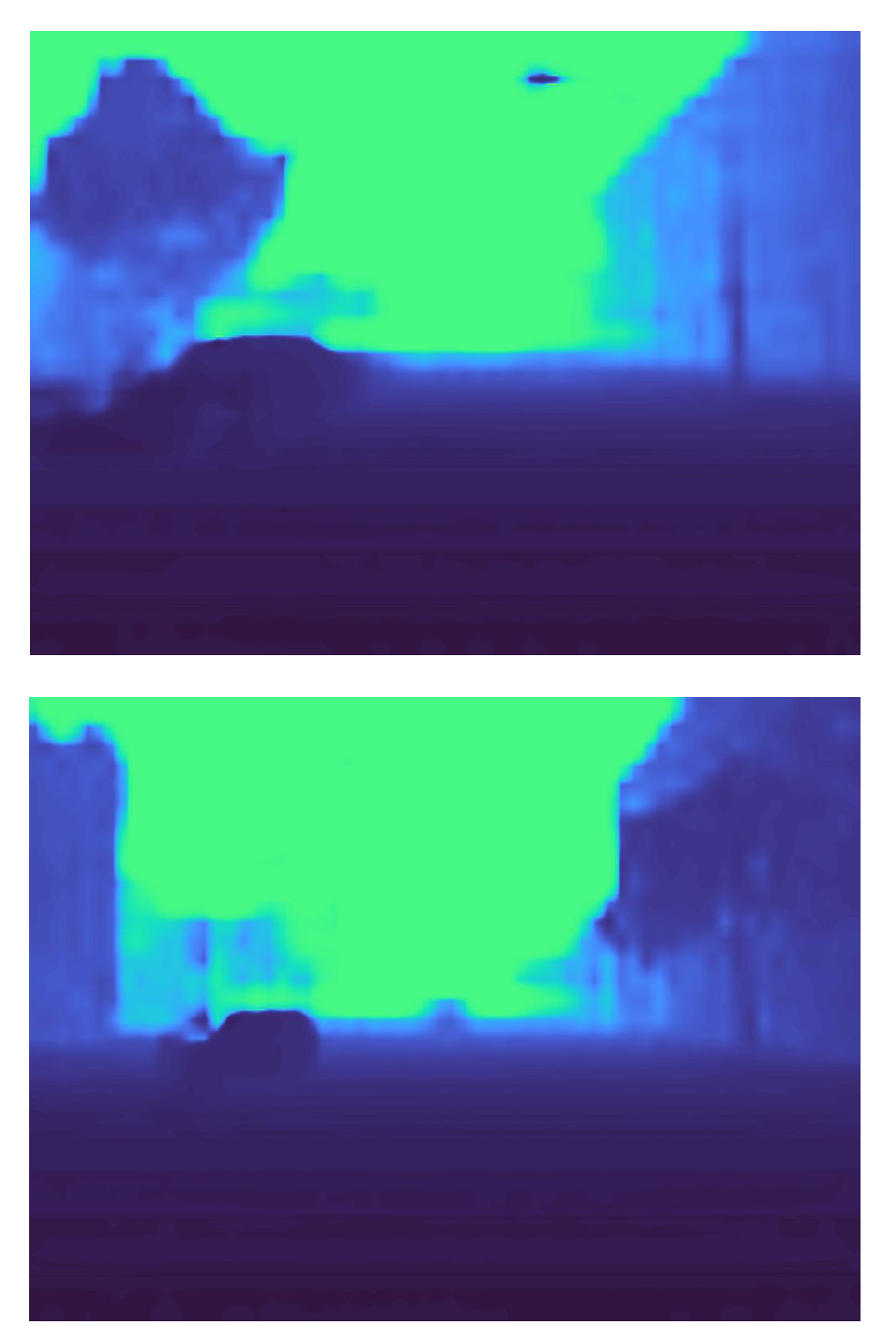}}     
	\subfigure[Fusion result]{\includegraphics[width=0.15\textwidth]{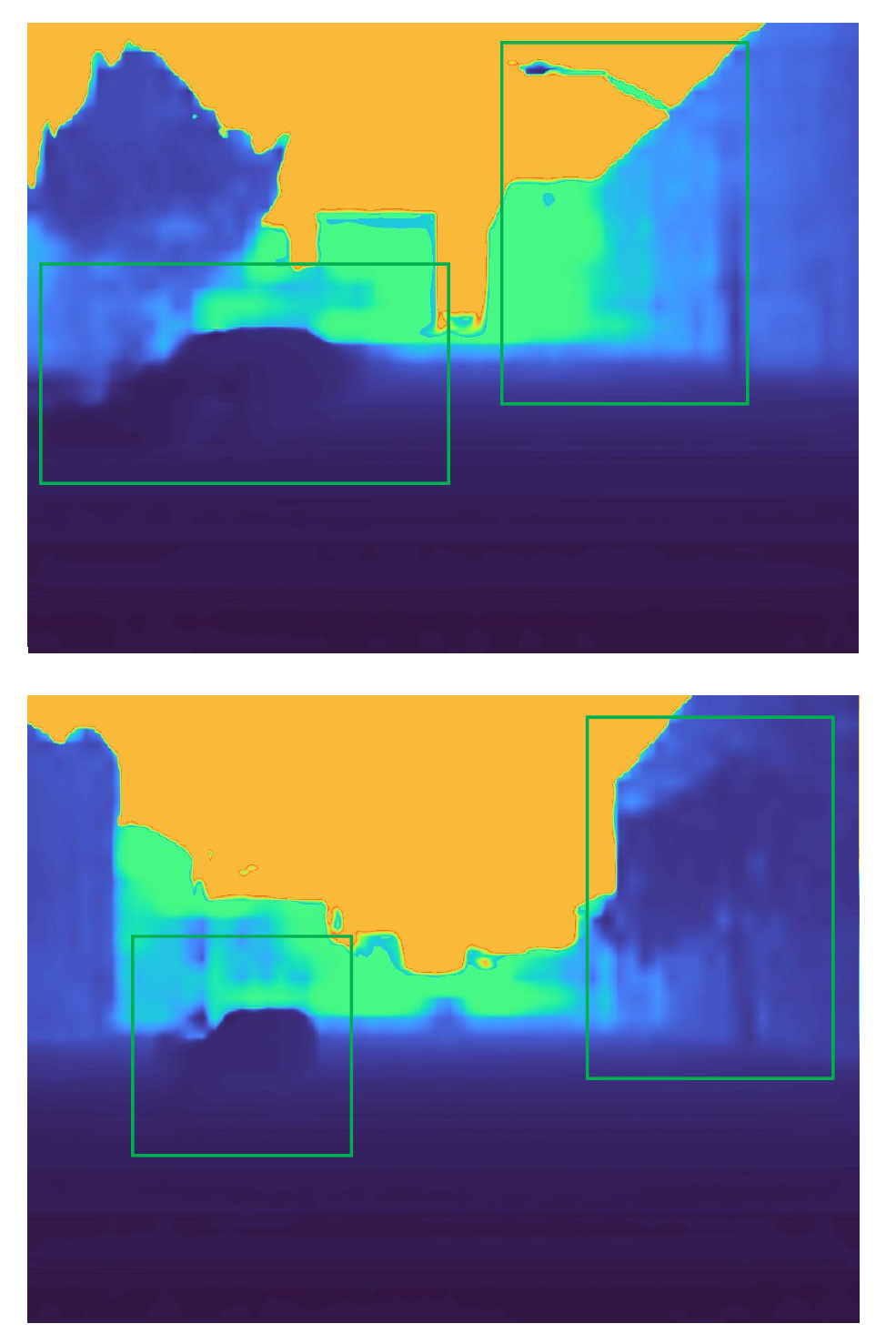}} 
	\subfigure[RGB result]{\includegraphics[width=0.15\textwidth]{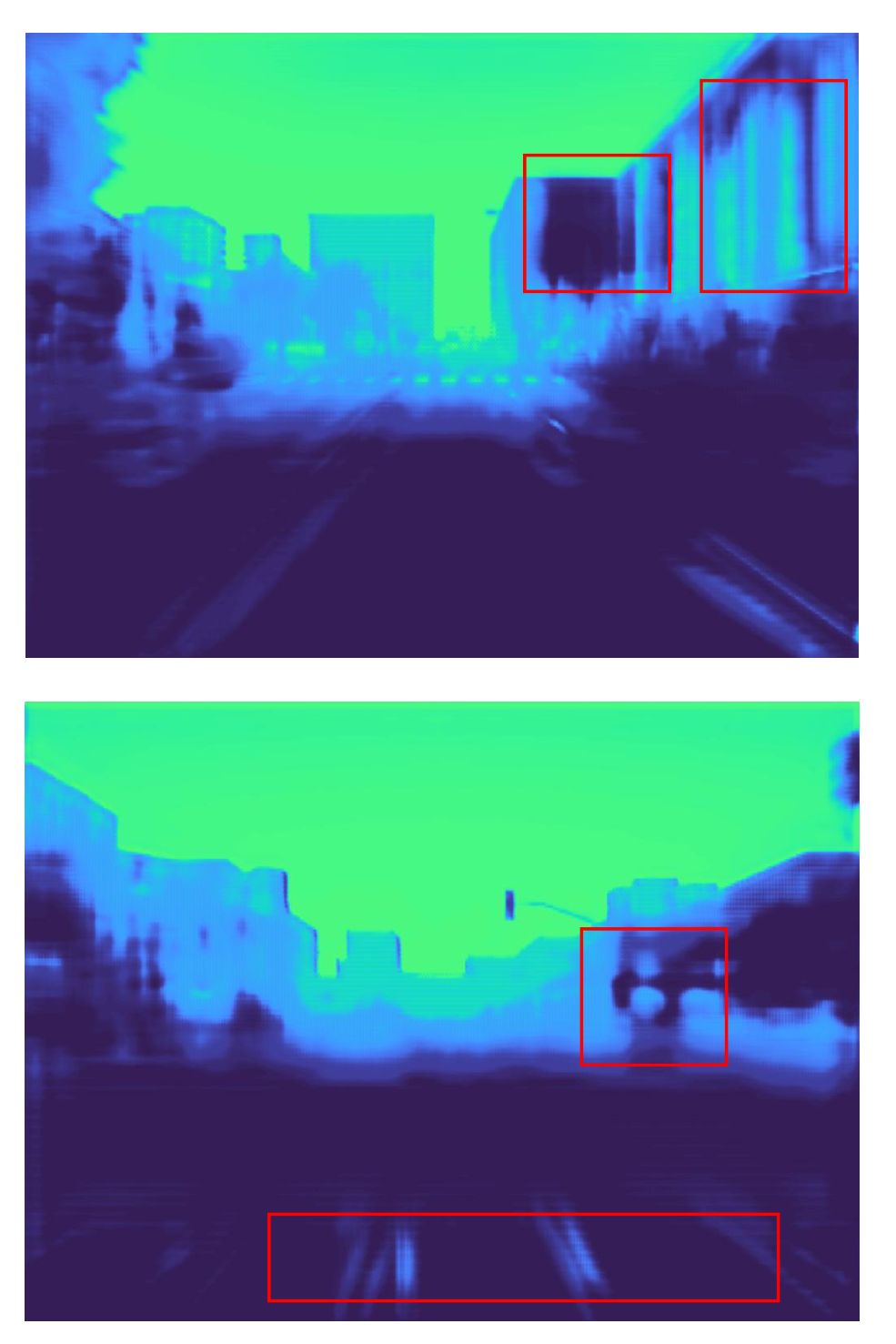}}\\
  
  \caption{Visualization of depth estimation on CitySpike20K. Pic. a is 30hz RGB data, and b is one spike frame in a spike voxel. Pic. c-e is output result of our method, and f is UNet output for RGB depth estimation. }
	\label{fig:data_distribution}
\end{figure*}

\textbf{Main Results and Analysis for CS20K} Table 1 and Table 2 show quantitative results with the comparison of the RGB methods. We experiment with classic monocular depth estimation works, as well as stereo depth estimation methods. We can see that under the uncertainty-guided fusion, our result gets the top performance among all the methods. Compared with the best monocular method, UGDF reduces $4.93\%$, $2.688$ error in terms of AbsRel.and RMSE metric respectively. For stereo methods, we also gained improvements on all metrics. We also show the qualitative comparison in Figure 5. As can be seen, our method achieves better depth estimation compared with blurred RGB-based methods. Other visualization results are in our supplement.



\begin{table*}[t]
  \centering
  \scriptsize
  \caption{Quantitative results on Spike-Real \textbf{test} set. Our UGDF framework still obtains performance increase to two branches.} 
    \begin{tabular}{c|c|c|c|ccccccc}
    \toprule

    Dataset & Method & Approach & Modality &  \multicolumn{1}{l}{Abs\_Rel↓}  \cellcolor{lightgray}  & \multicolumn{1}{l}{RMSE ↓} & \multicolumn{1}{l}{Sq\_Rel ↓} & \multicolumn{1}{l}{RMSE\_log ↓} & \multicolumn{1}{l}{a1 ↑} & \multicolumn{1}{l}{a2 ↑} & \multicolumn{1}{l}{a3 ↑} \\
    \midrule
    
    Real & PSMNet & Ster. & Image & 0.3743  & 2.228   &   0.413  &  0.843  & 0.451  & 0.703 & 0.838  \\     \midrule
    &  & Ster. &  & 0.2722  & 1.264  &   0.376  &  0.348  & 0.581  & 0.819 & 0.906\\  
   Real & \textbf{UGDF}(Ours)  & Mono. & Spike & 0.4037 & 1.552   &   1.017  &  0.382  & 0.528  & 0.796& 0.889 \\
   &  & Fusion &  & \textbf{0.2693}  & 1.237   &   0.413  &  0.374  & 0.533  & 0.795 & 0.899  \\
    \bottomrule
    \end{tabular}%
  \label{tab:1}%
\end{table*}%

\begin{table}[h]
  \centering
  \caption{Ablating the fusion effectiveness on CitySpike 20K. We design a depth estimation fusion method with strong-efficiency and gain improvements for both branches. } 
    \begin{tabular}{c|c|ccc}
    \toprule
    Split & Branch & \multicolumn{1}{l}{Abs\_Rel} & \multicolumn{1}{l}{Sq\_Rel } & a1   \\
    \midrule
   Valid  & Mono. & 0.3302  \textbf{(0.102↓)}& 12.759 & 0.738 \\

       & Ster. & 0.2543 (0.026↓) & 3.995 & 0.613  \\
 
    & E. Fusion & 0.2652 (0.037↓)& 5.712 & 0.706 \\    
    
    & U. Fusion & 0.2282 & 4.699 & 0.754 \\ 

    \midrule

  Test  & Mono. & 0.2944 \textbf{(0.095↓)} & 12.508 & 0.779 \\

        & Ster. & 0.2118 (0.012↓) & 3.780 & 0.753  \\

    & E. Fusion & 0.2347 (0.035↓) & 4.018 & 0.761 \\ 
    
    &U. Fusion. & 0.1997 & 4.897 & 0.791 \\ 
    
    \bottomrule
    \end{tabular}%
  \label{tab:1}%
\end{table}%
 
\begin{table}[h!]
  \begin{center}
  \small
    \caption{Ablation results on test split of CS20K for window-width of neuromorphic encoding. The runtime statistics are made on RTX 2080ti GPU for a single forward pass of the network with the batch-size of 1}
    \begin{tabular}{c|c|ccccc} 
      \toprule
      Width &  Time&   &  & \textbf{Error} & & \\\
       & (ms) & Abs\_Rel & Sq\_Rel &a1  &  a2 & a3 \\
      \midrule
      8 & 5.3 &0.2301 & 5.146 & 0.756 & 0.877 & 0.942 \\
      16 & 2.8 &0.2143 & 4.699 & 0.764 & 0.892 & 0.946 \\
      24 & 2.1 &0.1997 & 4.879 &  0.790 & 0.888 & 0.945 \\
      32 & 1.6 &0.2552 & 5.612  & 0.726 & 0.876  & 0.934 \\
      \bottomrule
    \end{tabular}
  \end{center}
  \label{tab3}
\end{table}

\textbf{Evaluation on Spike-Real Set}
We also train and evaluate our network on a dataset captured by a stereo real-world Vidar in a series of outdoor scenes. The dataset contains 40 sequences of outdoor scenes and we split 33 sequences for training and 7 sequences for testing. Table 4 and Figure 6 shows results evaluated on its test set.

\subsection{Ablation Study}
\label{sec:5.3}
We carry out ablation experiments from two aspects. The first of those is to explore the effect of different choices of time-window widths, and the other is to verify the effectiveness of uncertainty-guided fusion design. 


\textbf{Effectiveness of Uncertainty Guided Fusion}
We conduct experiments to verify the effectiveness of monocular and stereo uncertainty jointly guided fusion. In order to demonstrate the benefits of joint-guided fusion, we first compare it with a linear additive ensemble fusion manner. To be specific, we make a uniform linear addition of monocular and stereo estimation results, denoted as E. Fusion in Table 4 As can be seen, the linear additive fusion manner is inferior to other fusion methods. In addition, we visualize the improvement gap between fusion results and the other two branches. We can see the advantages of our UGDF framework. Firstly it combines both advantages of stereo and monocular estimation. And secondly, it brings substantial improvements, rather than the compromising fusion of ensemble style.

\textbf{Time window Width of Neuromorphic Encoding} 
In our framework, we apply a kind of neuromorphic encoding method to effectively extract the feature of spike data. As we have described in Section 3.2, we chunk the spike voxel into spike sequences by the time window of 24 to obtain better local representations. Then the sequences are sent into the Conv-RNN to extract temporal connections between different sequences. Theoretically, applying a smaller time window is beneficial to extract local connections between spike sequences, yet increases the convergence and inference time on hardware. So we set different widths of encoding time-window and train the whole network for 200 epochs. We make evaluations on the test set of CitySpike20K. The results are shown in Table 5, and we can see that 24 turns out to be the optimal choice of time-window widths for shorter inference time and better precision.

\section{Conclusion}
In this paper, we propose an uncertainty-guided depth fusion framework (UGDF) for spike data, consisting of four modules including neuromorphic encoding module, spike encoder, spike decoder for monocular and stereo tasks, and an uncertainty-guided fusion part. The main motivation of our work is monocular depth estimation models and stereo models show different advantages when performing predictions for spike data. So it's critical to explore an effective fusion method for leveraging the advantages of both tasks. Different from previous works, we fuse monocular and stereo depth prediction results according to individual adaptive uncertainty estimations. We also generate a spike dataset for depth estimation which contains 20K paired spike-depth data (CitySpike20K), along with its technical details and evaluation metrics. We demonstrate the good efficiency of spike data when applied to fast-moving circumstances. Extensive experiments are conducted to validate the effectiveness of our proposed UGDF. We hope this paper can inspire future works on spike data depth estimation.

\section*{Appendix I: Proposed Dataset: CitySpike20K}

\subsection*{Introduction and Visualization}
We propose CitySpike20K, a spike-depth dataset to help explore the depth estimation algorithms for spike camera. The dataset is generated by Unity3D and contains 10 sequences, 5 of which are day scenes and 5 others are night scenes. In the dataset, the frequency of the spike data and corresponding depth GTs is 1000Hz. Besides, we supply 30Hz RGB images for each scenes as well as 1000Hz RGB images that aligned with spike data.

To fully simulate the city environments, we add moving automobiles and dynamic traffic lights. We set 5-10 moving automobiles including buses, cars, vans and trucks for each scene. Figure 6 gives a visualization of CitySpike20K which contains RGB frames, spike data and depth maps. Specifically, we split scene03, scene07 for testing, scene09 for validation and others for training. 

\begin{figure*}[t]
\includegraphics[width=13cm]{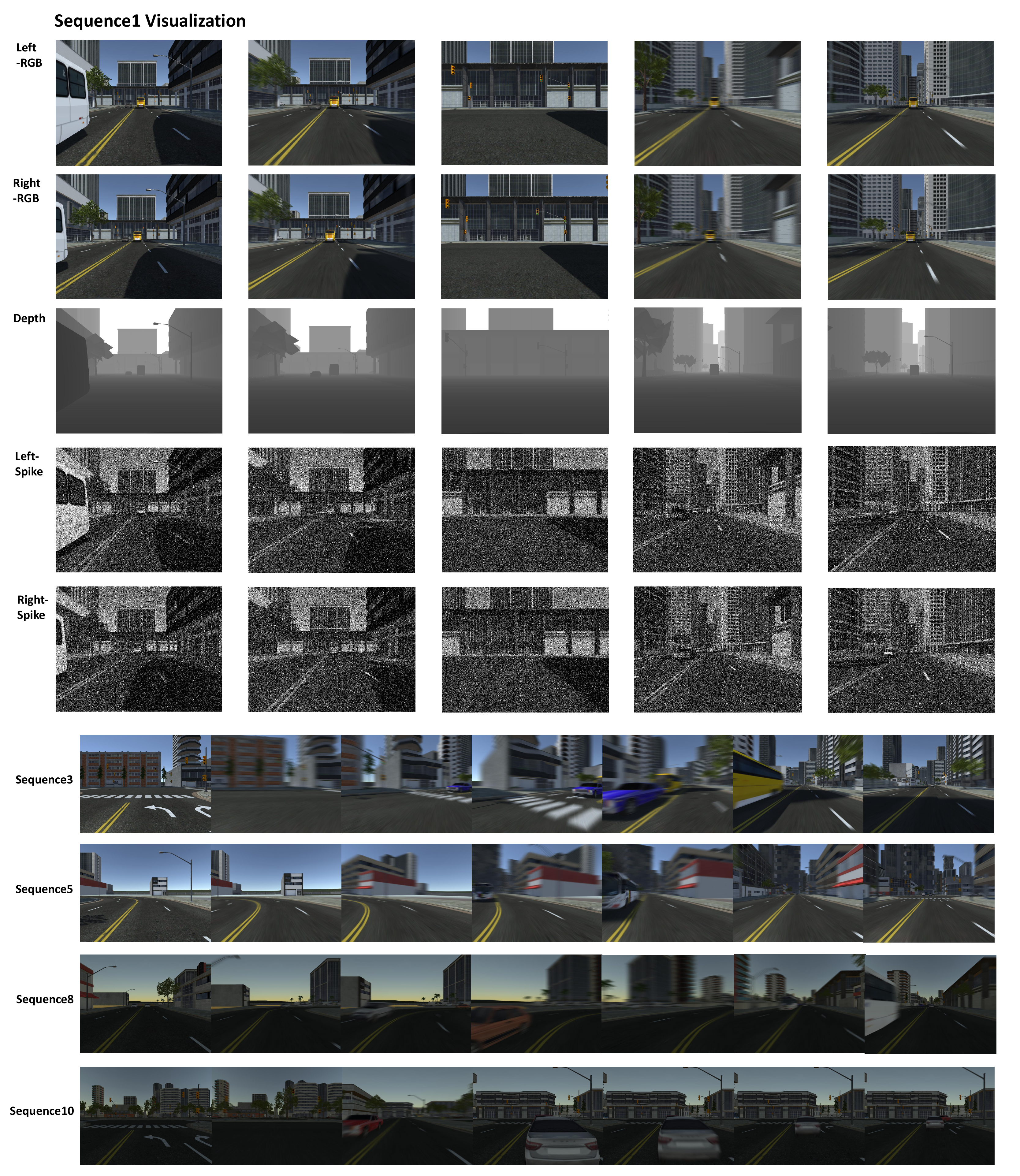}
\centering
\vspace{-5pt}
\caption{ A visualization of our proposed CitySpike20k dataset. We generate it by Unity3D engine and simulate a vivid city environment along with dense depth maps and spike data.}
\label{fig:2}
\end{figure*}

\begin{figure*}[t]
\includegraphics[width=16.5cm]{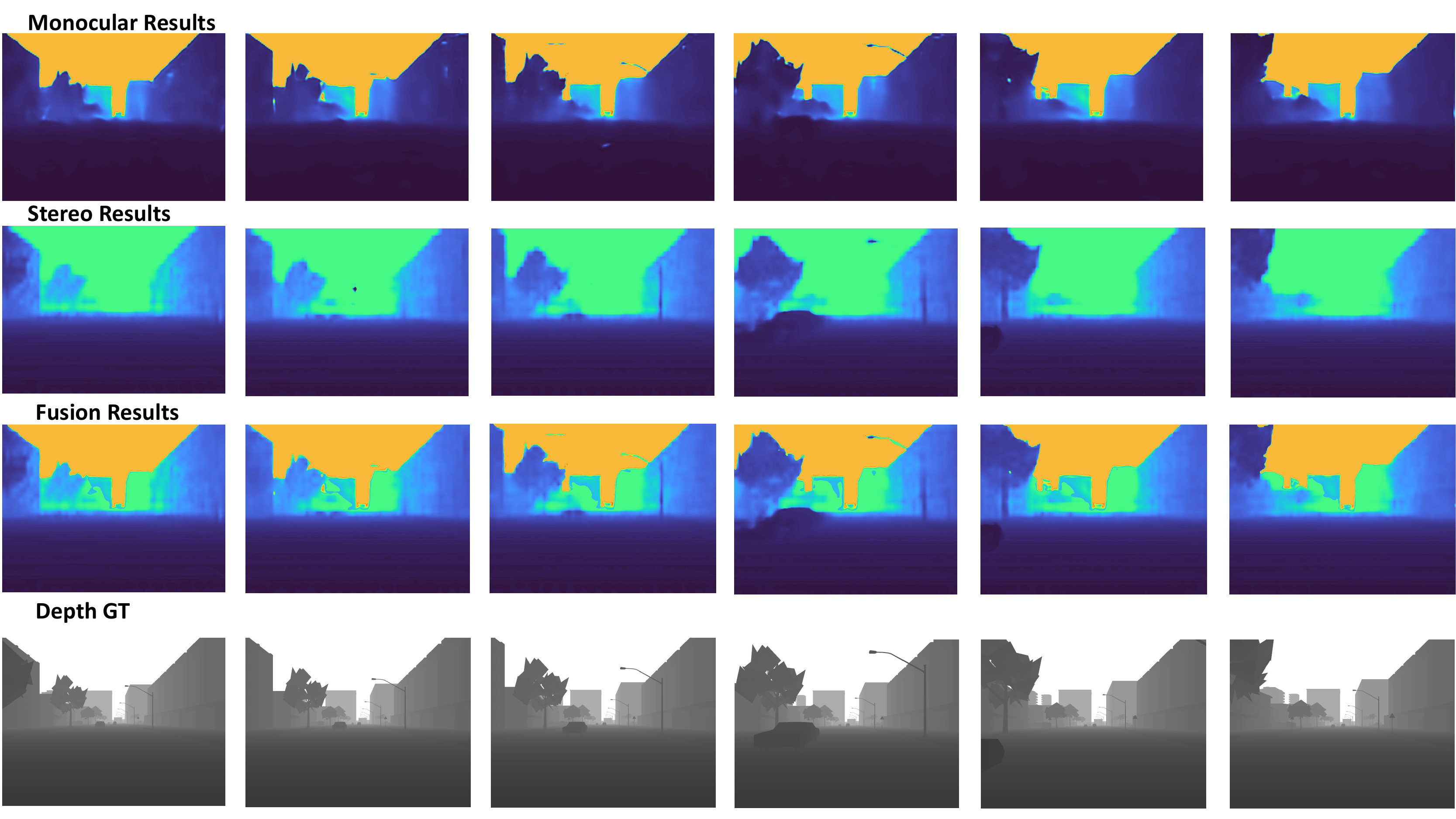}
\centering
\vspace{-5pt}
\caption{More prediction results on CitySpike20K dataset. As can be seen, the stereo estimation results and the monocular estimation results fuse efficiently by our framework}
\label{fig:2}
\end{figure*}

\begin{figure}[h]
\includegraphics[width=8cm]{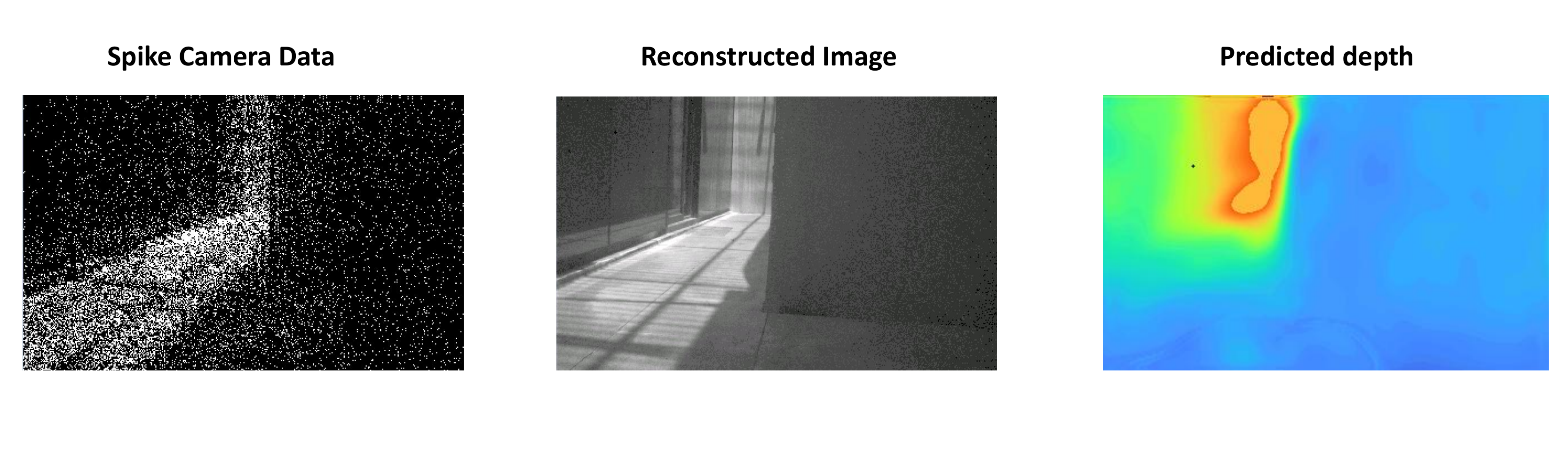}
\centering
\vspace{-5pt}
\caption{A visualization for Spike-Real dataset and prediction results from its test set.}
\end{figure}

\begin{table*}[t]
  \centering
  \footnotesize
  \caption{Quantitative results on Spike-Kitti \textbf{test} set. Our UGDF framework still obtains performance increase to two branches.} 
    \begin{tabular}{c|c|c|c|ccccccc}
    \toprule
    Dataset & Method & Approach & Modality &   \multicolumn{1}{l}{Abs\_Rel↓}  \cellcolor{lightgray} & \multicolumn{1}{l}{RMSE ↓} & \multicolumn{1}{l}{Sq\_Rel ↓} & \multicolumn{1}{l}{RMSE\_log ↓} & \multicolumn{1}{l}{a1 ↑} & \multicolumn{1}{l}{a2 ↑} & \multicolumn{1}{l}{a3 ↑} \\
    \midrule
    &  & Ster. &  & 0.1250 & 4.283 &   0.717  &  0.188  & 0.830  & 0.957 & 0.986\\  
   Spike-Kitti& \textbf{UGDF}(Ours)  & Mono. & Spike & 0.1706 & 5.067 &   1.127  &  0.242  & 0.753  & 0.910& 0.968 \\
   &  & Fusion &  & \textbf{0.1247}  & 4.281   &   0.721  &  0.189  & 0.829  & 0.957 & 0.985  \\
    \bottomrule
    \end{tabular}%
  \label{tab:1}%
\end{table*}%

\begin{table*}[t]
  \centering
  \footnotesize
  \caption{Quantitative results on CitySpike20K-demo. Evaluation metrics are as described above. We make comparison with DORN\cite{fu2018deep}, GwcNet\cite{guo2019group}, CFNet\cite{shen2021cfnet}, StereoNet\cite{Khamis_2018_ECCV}, PSMNet\cite{chang2018pyramid}, and GANet\cite{Zhang_2019_CVPR} . The evaluation metrics are as introduced in subsection 4.2. We also consider model parameter size to be one of compared targets.} 
    
    \begin{tabular}{c|c|c|ccccccc}
    \toprule
    Dataset & Method & Approach & \multicolumn{1}{l}{Abs\_Rel↓} & \multicolumn{1}{l}{RMSE ↓} & \multicolumn{1}{l}{Sq\_Rel ↓} & \multicolumn{1}{l}{RMSE\_log ↓} & \multicolumn{1}{l}{a1 ↑} & \multicolumn{1}{l}{a2 ↑} & \multicolumn{1}{l}{a3 ↑} \\
    \midrule
     & UNet\cite{ronneberger2015u}  & Mono.  & 0.2518 & \underline{23.993} & 9.008 & 0.357 & 0.68  & \underline{0.896} & 0.932 \\
    
   demo & DORN\cite{fu2018deep}  & Mono. & 0.3857 & 25.258 & 10.691 & 0.438 & 0.409 & 0.841& 0.917\\
    & Eigen\cite{eigen2014depth}  & Mono.& 0.4262 & 25.154 & 20.363 & 0.459 & 0.542 & 0.800& 0.893 \\
    \midrule
    & GC-Net\cite{kendall2017end} & Ster.  &0.2350& 37.158 & 12.743 & 0.401 & 0.614 & 0.809 & 0.868\\    
    & GwcNet\cite{guo2019group} & Ster. & \underline{0.1880} & 24.152 & 7.469 & \textbf{0.304} & \underline{0.757} & 0.895& \underline{0.953} \\
   
    & CFnet\cite{shen2021cfnet} & Ster. &  0.2281 & 25.905 & \textbf{5.557} & 0.397 & 0.610  & 0.847&0.926 \\
    
  demo & SteroNet\cite{Khamis_2018_ECCV} & Ster.& 0.2890 & 50.765 & 19.772 & 0.690 & 0.563& 0.727& 0.823\\
    
    & PSMNet\cite{chang2018pyramid} & Ster. & 0.1886 & 28.496 & \underline{7.354} & 0.340& 0.723 & 0.887 &0.941 \\
    & GANet-1\cite{Zhang_2019_CVPR} & Ster.&0.3270 & 49.068 & 19.505 & 0.865 & 0.586 & 0.764 & 0.851 \\
    & GANet\cite{Zhang_2019_CVPR} & Ster.&0.2963 & 47.202 & 17.598 & 0.714 & 0.576 & 0.771 & 0.857 \\  
    \midrule
   demo & \textbf{Ours } & Fusion & \textbf{0.1715 } &   \textbf{22.793 }   &   11.217 &   \underline{0.306}   &   \textbf{0.791}    & \textbf{0.928}&\textbf{0.961} \\
    \bottomrule
    \end{tabular}%
  \label{tab:1}%
\end{table*}%

\begin{figure*}[h]
\includegraphics[width=16.5cm]{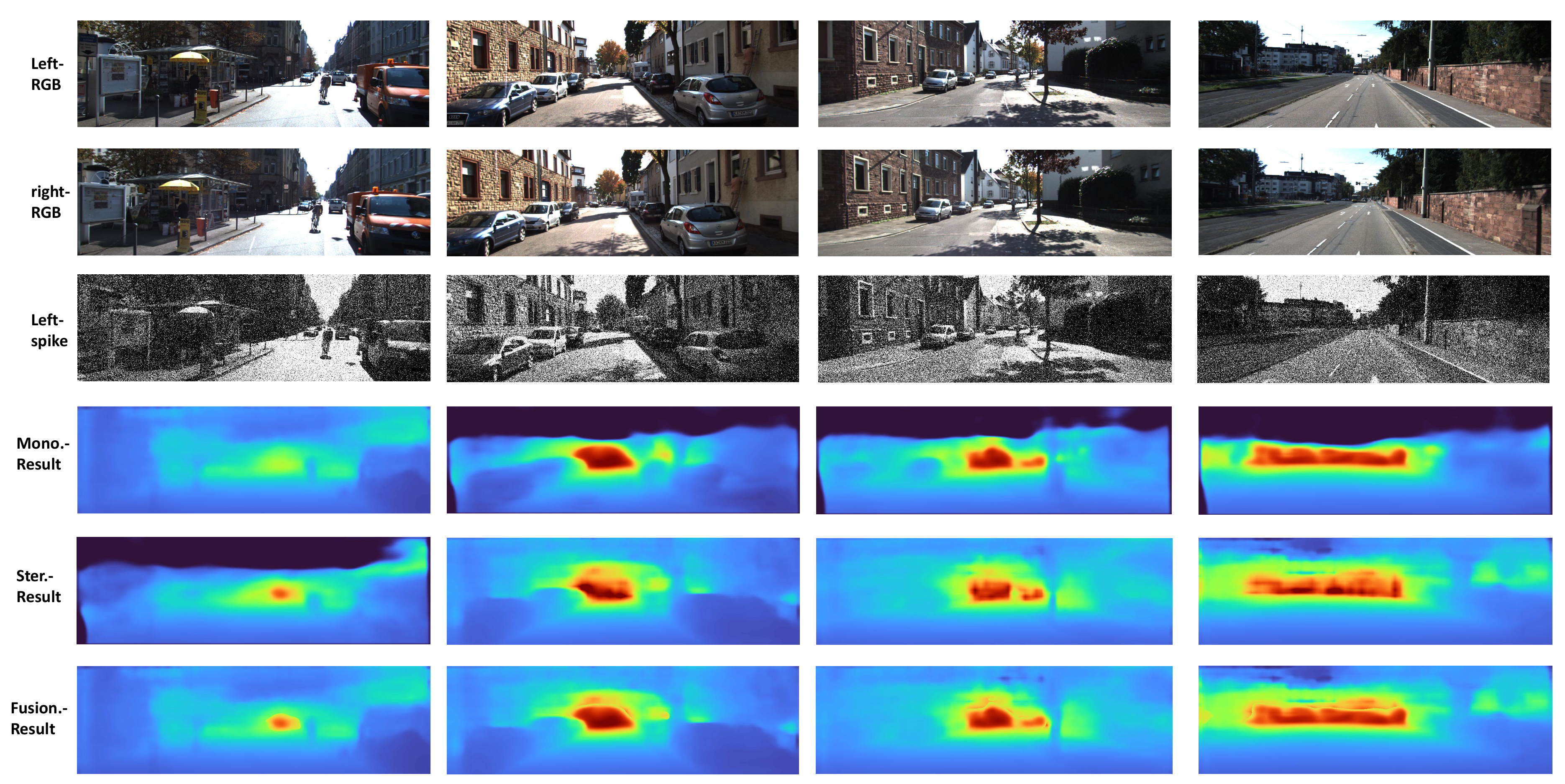}
\centering
\vspace{-5pt}
\caption{Visualization of predicting results on Spike-Kitti datasets. As seen, the monocular branch can still provide smoother and more accurate results in further regions, and the stereo branch makes sharper and cleaner prediction for closer regions.}

\end{figure*}

\subsection*{Evaluation Metric}
We conducted to evaluate the effectiveness of supervised depth estimation model on CitySpike20K. Our evaluation metrics for depth estimation is described as follows:

Given an estimated depth map $\hat{D}$, and its corresponding ground truth $D$, $N = H \times W$, $Abs\_Rel$ is quantified as:
\begin{equation}
    Abs\_Rel = \frac{1}{N}\sum_{i=1}^{N}\frac{|D_i - \hat{D_i}|}{D_i}
\end{equation}
and RMSE defined:
\begin{equation}
    RMSE = \sqrt{\frac{1}{N}\sum_{i=1}^{N}||D_i - \hat{D_i}||^{2}}
\end{equation}
we also introduce $RMSE\_log$ metric:
\begin{equation}
    RMSE\_log = \sqrt{\frac{1}{N}\sum_{i=1}^{N}||log(\hat{D_i})-log(D_i)||^{2}}
\end{equation}
and Sq\_Rel metric as here:
\begin{equation}
    Sq\_Rel = \frac{1}{N}\sum_{i=1}^{N}\frac{||D_i - \hat{D_i}||^{2}}{D_i}
\end{equation}
Above metrics measure output errors from different statistic aspect, weighting the distance between predictions and ground-truth labels, where lower values mean better model performance. Below metrics are for evaluation of whether predictions are accurate within certain range of ground-truth, and higher values mean better performance. Note that $j\in \{1,2,3\}$

\begin{equation}
    aj \quad accuracy: \% \quad of \quad D_i \quad s.t. \quad max(\frac{\hat{D_i}}{D_i},\frac{D_i}{\hat{D_i}})=\delta<T=1.25^{j}
\end{equation}

\begin{figure*}[t]
\includegraphics[width=13.0cm]{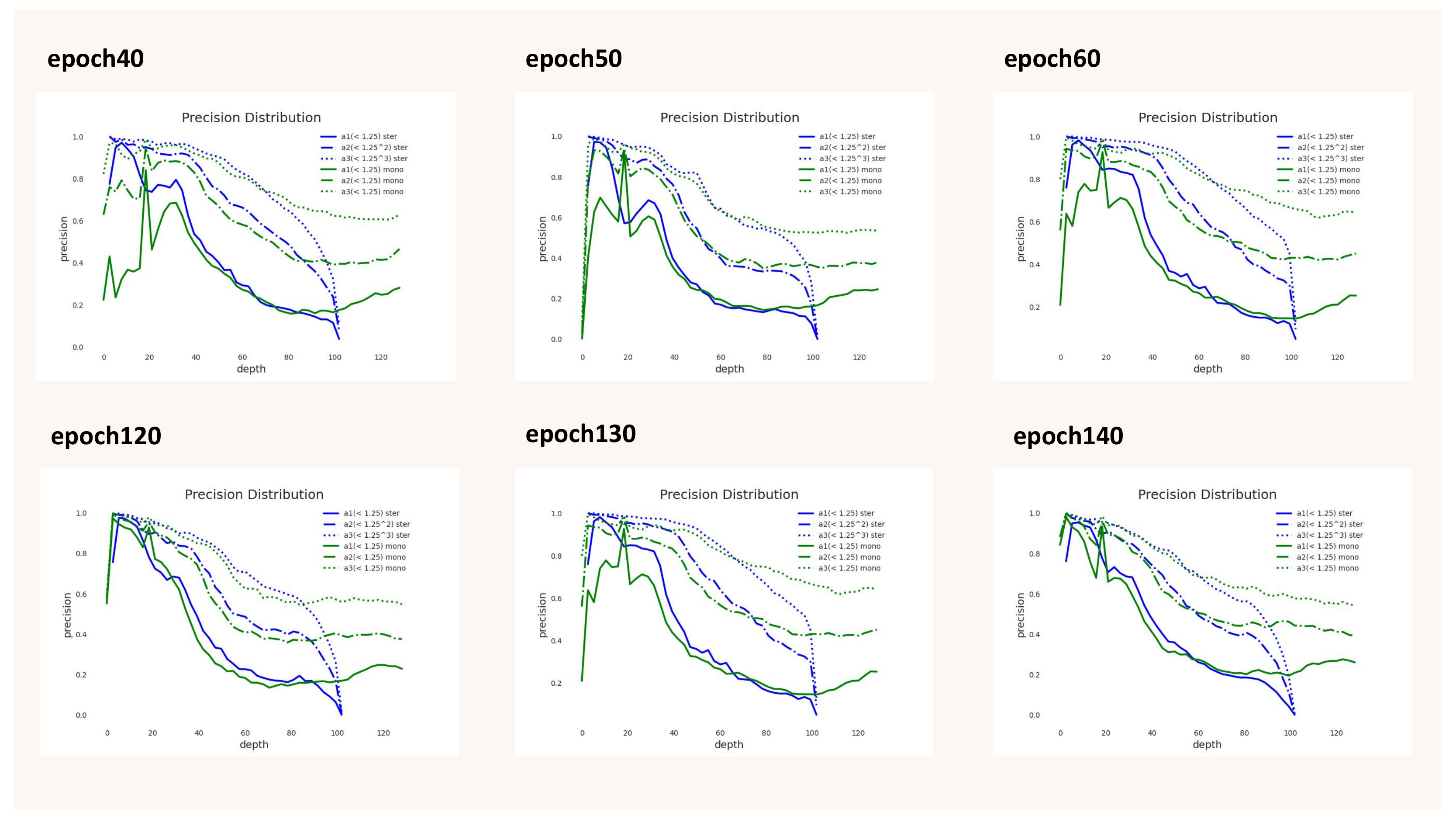}
\centering

\caption{Accuracy statistics on CitySpike20K test set. The green lines and blue lines represent the monocular and stereo accuracies respectively.}
\label{fig:2}
\end{figure*}

\begin{figure*}[h]
\includegraphics[width=13.0cm]{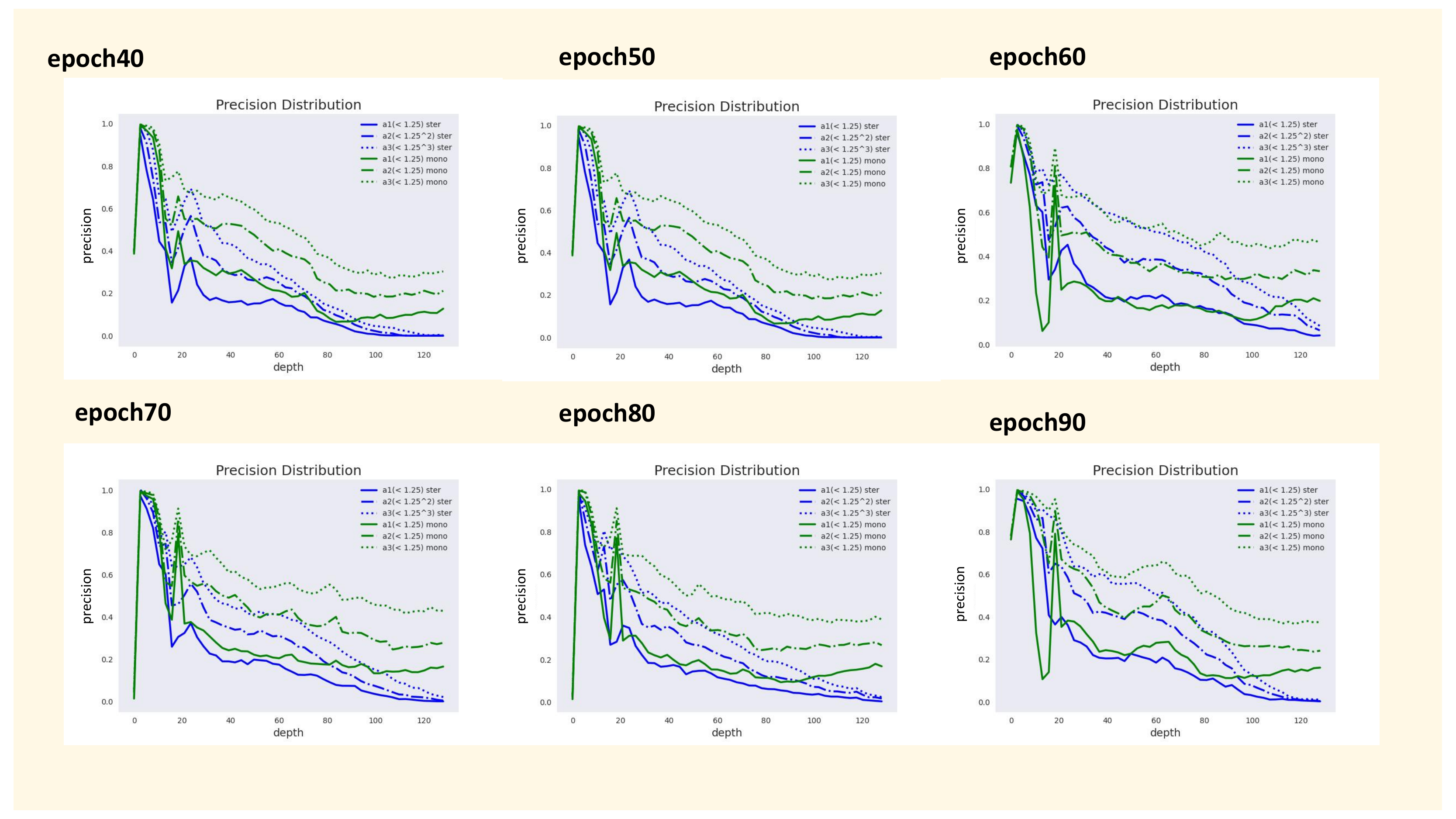}
\centering

\caption{Accuracy statistics on CitySpike20K validation set. }
\label{fig:2}
\end{figure*}

\section*{Appendix II: Performance on other datasets}

\subsection*{Real-Dataset}
As we have described in our submitted paper, we also evaluate our framework on a real-recorded dataset by a spike camera. The dataset contains 40 sequences data and each of which includes 3-6 $ [400\times250\times400] $ spike voxels in the format of  $[T\times H \times W]$. We split 33 sequences for training and 7 for testing.

\subsection*{Kitti}
To demonstrate that our UGDF framework still works in real-world scenes, we carry out experiment on a spike-kitti dataset. To convert Kitti\cite{geiger2013vision} from RGB modality to spike modality, we first make frame interpolation using XVFI\cite{sim2021xvfi} by 128 times. Then we use a Simulated-Vidar code script to generate spike data from RGB Kitti images to form spike voxels in the format $(128\times 375\times 1242)$ , where 128 represents the time dimension and $(375\times1242)$ is the original size of Kitti RGB images. We maintain the same way to operate neuromorphic encoding as what we design for CitySpike20K dataset in our submitted paper. As mentioned above, we set this experiment to further explore the effectiveness of our fusion strategy. We train our framework for 50 epochs on 4 RTX-2080Ti GPUs. Figure 9 provides a group of visualization of the output of our results on validation set.

\subsection*{CitySpike20K-demo}
In addition to 10 sequences of 1000Hz spike data we provide in the CitySpike20K dataset, we still supply a 40000Hz demo to simulate real spike as possible as we could. The demo contains 60K paired data and records a 1.5 seconds video of a fast-driving car in the city street. Different from our submitted papers, we use this demo to evaluate the performance of models loaded with spike data. Considering existing methods for monocular or stereo depth estimation are mostly based on RGB 3-channel data, we change the input channel of the models to the time-window width of applied spike sequences, i.e. 32 as we adopted. And we use the first half of the demo for training and the second half for testing. 

\section*{Appendix III: Statistics to Support Our Motivation }
There are two clues to inspire our motivations. The first of which is that, the spike camera has its unique advantages to deal with fast-moving circumstances when operating depth estimation task. And the second is that, the monocular strategy and stereo strategy share some distinct advantages to finish depth estimation task while loaded with spike data. We supply statistical results to prove our second motivation. On CitySpike20K dataset, we make a1, a2, a3 accuracy calculation in different depth intervals according to depth GT while evaluating our network. We transform the stereo disparities into depths, and count a1, a2, a3 accuracy for two branches respectively in the same metrics. Then we plot them in one coordinate. Figure 10 shows statistical results on test set and Figure 11 shows results on validation sets. As seen, the stereo branch suffers from great accuracy decrease for far regions, while monocular branch still maintains certain reliability. Similarly, the stereo branch is more stable and accurate than the monocular branch for closer regions.

\bibliographystyle{unsrt}  
\bibliography{references}

\begin{thebibliography}{10}

\bibitem{tremblay2018deep}
Jonathan Tremblay, Thang To, Balakumar Sundaralingam, Yu~Xiang, Dieter Fox, and
  Stan Birchfield.
\newblock Deep object pose estimation for semantic robotic grasping of
  household objects.
\newblock {\em arXiv preprint arXiv:1809.10790}, 2018.

\bibitem{tang20193d}
Fulin Tang, Yihong Wu, Xiaohui Hou, and Haibin Ling.
\newblock 3d mapping and 6d pose computation for real time augmented reality on
  cylindrical objects.
\newblock {\em IEEE Transactions on Circuits and Systems for Video Technology},
  30(9):2887--2899, 2019.

\bibitem{marchand2015pose}
Eric Marchand, Hideaki Uchiyama, and Fabien Spindler.
\newblock Pose estimation for augmented reality: a hands-on survey.
\newblock {\em IEEE transactions on visualization and computer graphics},
  22(12):2633--2651, 2015.

\bibitem{manhardt2019roi}
Fabian Manhardt, Wadim Kehl, and Adrien Gaidon.
\newblock Roi-10d: Monocular lifting of 2d detection to 6d pose and metric
  shape.
\newblock In {\em Proceedings of the IEEE/CVF Conference on Computer Vision and
  Pattern Recognition}, pages 2069--2078, 2019.

\bibitem{wu20196d}
Di~Wu, Zhaoyong Zhuang, Canqun Xiang, Wenbin Zou, and Xia Li.
\newblock 6d-vnet: End-to-end 6-dof vehicle pose estimation from monocular rgb
  images.
\newblock In {\em Proceedings of the IEEE/CVF Conference on Computer Vision and
  Pattern Recognition Workshops}, pages 0--0, 2019.

\bibitem{hu2021optical}
Liwen Hu, Rui Zhao, Ziluo Ding, Lei Ma, Boxin Shi, Ruiqin Xiong, and Tiejun
  Huang.
\newblock Optical flow estimation for spiking camera.
\newblock {\em arXiv preprint arXiv:2110.03916}, 2021.

\bibitem{dong2019efficient}
Siwei Dong, Lin Zhu, Daoyuan Xu, Yonghong Tian, and Tiejun Huang.
\newblock An efficient coding method for spike camera using inter-spike
  intervals.
\newblock {\em arXiv preprint arXiv:1912.09669}, 2019.

\bibitem{zhu2019retina}
Lin Zhu, Siwei Dong, Tiejun Huang, and Yonghong Tian.
\newblock A retina-inspired sampling method for visual texture reconstruction.
\newblock In {\em 2019 IEEE International Conference on Multimedia and Expo
  (ICME)}, pages 1432--1437. IEEE, 2019.

\bibitem{zhu2020retina}
Lin Zhu, Siwei Dong, Jianing Li, Tiejun Huang, and Yonghong Tian.
\newblock Retina-like visual image reconstruction via spiking neural model.
\newblock In {\em Proceedings of the IEEE/CVF Conference on Computer Vision and
  Pattern Recognition}, pages 1438--1446, 2020.

\bibitem{mayer2016large}
Nikolaus Mayer, Eddy Ilg, Philip Hausser, Philipp Fischer, Daniel Cremers,
  Alexey Dosovitskiy, and Thomas Brox.
\newblock A large dataset to train convolutional networks for disparity,
  optical flow, and scene flow estimation.
\newblock In {\em Proceedings of the IEEE conference on computer vision and
  pattern recognition}, pages 4040--4048, 2016.

\bibitem{kendall2017end}
Alex Kendall, Hayk Martirosyan, Saumitro Dasgupta, Peter Henry, Ryan Kennedy,
  Abraham Bachrach, and Adam Bry.
\newblock End-to-end learning of geometry and context for deep stereo
  regression.
\newblock In {\em Proceedings of the IEEE international conference on computer
  vision}, pages 66--75, 2017.

\bibitem{Khamis_2018_ECCV}
Sameh Khamis, Sean Fanello, Christoph Rhemann, Adarsh Kowdle, Julien Valentin,
  and Shahram Izadi.
\newblock Stereonet: Guided hierarchical refinement for real-time edge-aware
  depth prediction.
\newblock In {\em Proceedings of the European Conference on Computer Vision
  (ECCV)}, September 2018.

\bibitem{Chabra_2019_CVPR}
Rohan Chabra, Julian Straub, Christopher Sweeney, Richard Newcombe, and Henry
  Fuchs.
\newblock Stereodrnet: Dilated residual stereonet.
\newblock In {\em Proceedings of the IEEE/CVF Conference on Computer Vision and
  Pattern Recognition (CVPR)}, June 2019.

\bibitem{guo2019group}
Xiaoyang Guo, Kai Yang, Wukui Yang, Xiaogang Wang, and Hongsheng Li.
\newblock Group-wise correlation stereo network.
\newblock In {\em Proceedings of the IEEE/CVF Conference on Computer Vision and
  Pattern Recognition}, pages 3273--3282, 2019.

\bibitem{Xu_2018_CVPR}
Dan Xu, Wei Wang, Hao Tang, Hong Liu, Nicu Sebe, and Elisa Ricci.
\newblock Structured attention guided convolutional neural fields for monocular
  depth estimation.
\newblock In {\em Proceedings of the IEEE Conference on Computer Vision and
  Pattern Recognition (CVPR)}, June 2018.

\bibitem{Ramamonjisoa_2020_CVPR}
Michael Ramamonjisoa, Yuming Du, and Vincent Lepetit.
\newblock Predicting sharp and accurate occlusion boundaries in monocular depth
  estimation using displacement fields.
\newblock In {\em Proceedings of the IEEE/CVF Conference on Computer Vision and
  Pattern Recognition (CVPR)}, June 2020.

\bibitem{lee2019monocular}
Jae-Han Lee and Chang-Su Kim.
\newblock Monocular depth estimation using relative depth maps.
\newblock In {\em Proceedings of the IEEE/CVF Conference on Computer Vision and
  Pattern Recognition}, pages 9729--9738, 2019.

\bibitem{Ramamonjisoa_2019_ICCV}
Michael Ramamonjisoa and Vincent Lepetit.
\newblock Sharpnet: Fast and accurate recovery of occluding contours in
  monocular depth estimation.
\newblock In {\em Proceedings of the IEEE/CVF International Conference on
  Computer Vision (ICCV) Workshops}, Oct 2019.

\bibitem{fu2018deep}
Huan Fu, Mingming Gong, Chaohui Wang, Kayhan Batmanghelich, and Dacheng Tao.
\newblock Deep ordinal regression network for monocular depth estimation.
\newblock In {\em Proceedings of the IEEE conference on computer vision and
  pattern recognition}, pages 2002--2011, 2018.

\bibitem{godard2017unsupervised}
Cl{\'e}ment Godard, Oisin Mac~Aodha, and Gabriel~J Brostow.
\newblock Unsupervised monocular depth estimation with left-right consistency.
\newblock In {\em Proceedings of the IEEE conference on computer vision and
  pattern recognition}, pages 270--279, 2017.

\bibitem{liu2022local}
Biyang Liu, Huimin Yu, and Yangqi Long.
\newblock Local similarity pattern and cost self-reassembling for deep stereo
  matching networks.
\newblock In {\em Proceedings of the AAAI Conference on Artificial
  Intelligence}, volume~36, pages 1647--1655, 2022.

\bibitem{chang2018pyramid}
Jia-Ren Chang and Yong-Sheng Chen.
\newblock Pyramid stereo matching network.
\newblock In {\em Proceedings of the IEEE conference on computer vision and
  pattern recognition}, pages 5410--5418, 2018.

\bibitem{chen2021revealing}
Zhi Chen, Xiaoqing Ye, Wei Yang, Zhenbo Xu, Xiao Tan, Zhikang Zou, Errui Ding,
  Xinming Zhang, and Liusheng Huang.
\newblock Revealing the reciprocal relations between self-supervised stereo and
  monocular depth estimation.
\newblock In {\em Proceedings of the IEEE/CVF International Conference on
  Computer Vision}, pages 15529--15538, 2021.

\bibitem{zhou2021sub}
Hang Zhou, Sarah Taylor, and David Greenwood.
\newblock Sub-depth: Self-distillation and uncertainty boosting self-supervised
  monocular depth estimation.
\newblock {\em arXiv preprint arXiv:2111.09692}, 2021.

\bibitem{eigen2014depth}
David Eigen, Christian Puhrsch, and Rob Fergus.
\newblock Depth map prediction from a single image using a multi-scale deep
  network.
\newblock {\em Advances in neural information processing systems}, 27, 2014.

\bibitem{ranftl2021vision}
Ren{\'e} Ranftl, Alexey Bochkovskiy, and Vladlen Koltun.
\newblock Vision transformers for dense prediction.
\newblock In {\em Proceedings of the IEEE/CVF International Conference on
  Computer Vision}, pages 12179--12188, 2021.

\bibitem{yang2021transformer}
Guanglei Yang, Hao Tang, Mingli Ding, Nicu Sebe, and Elisa Ricci.
\newblock Transformer-based attention networks for continuous pixel-wise
  prediction.
\newblock In {\em Proceedings of the IEEE/CVF International Conference on
  Computer Vision}, pages 16269--16279, 2021.

\bibitem{guizilini20203d}
Vitor Guizilini, Rares Ambrus, Sudeep Pillai, Allan Raventos, and Adrien
  Gaidon.
\newblock 3d packing for self-supervised monocular depth estimation.
\newblock In {\em Proceedings of the IEEE/CVF Conference on Computer Vision and
  Pattern Recognition}, pages 2485--2494, 2020.

\bibitem{lyu2020hr}
Xiaoyang Lyu, Liang Liu, Mengmeng Wang, Xin Kong, Lina Liu, Yong Liu, Xinxin
  Chen, and Yi~Yuan.
\newblock Hr-depth: High resolution self-supervised monocular depth estimation.
\newblock {\em arXiv preprint arXiv:2012.07356}, 6, 2020.

\bibitem{bhat2021adabins}
Shariq~Farooq Bhat, Ibraheem Alhashim, and Peter Wonka.
\newblock Adabins: Depth estimation using adaptive bins.
\newblock In {\em Proceedings of the IEEE/CVF Conference on Computer Vision and
  Pattern Recognition}, pages 4009--4018, 2021.

\bibitem{roy2016monocular}
Anirban Roy and Sinisa Todorovic.
\newblock Monocular depth estimation using neural regression forest.
\newblock In {\em Proceedings of the IEEE conference on computer vision and
  pattern recognition}, pages 5506--5514, 2016.

\bibitem{Jiao_2018_ECCV}
Jianbo Jiao, Ying Cao, Yibing Song, and Rynson Lau.
\newblock Look deeper into depth: Monocular depth estimation with semantic
  booster and attention-driven loss.
\newblock In {\em Proceedings of the European Conference on Computer Vision
  (ECCV)}, September 2018.

\bibitem{hoyer2021three}
Lukas Hoyer, Dengxin Dai, Yuhua Chen, Adrian Koring, Suman Saha, and Luc
  Van~Gool.
\newblock Three ways to improve semantic segmentation with self-supervised
  depth estimation.
\newblock In {\em Proceedings of the IEEE/CVF Conference on Computer Vision and
  Pattern Recognition}, pages 11130--11140, 2021.

\bibitem{4270415}
Kristian Ambrosch, Wilfried Kubinger, Martin Humenberger, and Andreas
  Steininger.
\newblock Hardware implementation of an sad based stereo vision algorithm.
\newblock In {\em 2007 IEEE Conference on Computer Vision and Pattern
  Recognition}, pages 1--6, 2007.

\bibitem{7989227}
Kai Berger, Randolph Voorhies, and Larry~H. Matthies.
\newblock Depth from stereo polarization in specular scenes for urban robotics.
\newblock In {\em 2017 IEEE International Conference on Robotics and Automation
  (ICRA)}, pages 1966--1973, 2017.

\bibitem{Zhou_2017_ICCV}
Chao Zhou, Hong Zhang, Xiaoyong Shen, and Jiaya Jia.
\newblock Unsupervised learning of stereo matching.
\newblock In {\em Proceedings of the IEEE International Conference on Computer
  Vision (ICCV)}, Oct 2017.

\bibitem{Tosi_2019_CVPR}
Fabio Tosi, Filippo Aleotti, Matteo Poggi, and Stefano Mattoccia.
\newblock Learning monocular depth estimation infusing traditional stereo
  knowledge.
\newblock In {\em Proceedings of the IEEE/CVF Conference on Computer Vision and
  Pattern Recognition (CVPR)}, June 2019.

\bibitem{zhu2018multivehicle}
Alex~Zihao Zhu, Dinesh Thakur, Tolga {\"O}zaslan, Bernd Pfrommer, Vijay Kumar,
  and Kostas Daniilidis.
\newblock The multivehicle stereo event camera dataset: An event camera dataset
  for 3d perception.
\newblock {\em IEEE Robotics and Automation Letters}, 3(3):2032--2039, 2018.

\bibitem{zhu2018realtime}
Alex~Zihao Zhu, Yibo Chen, and Kostas Daniilidis.
\newblock Realtime time synchronized event-based stereo.
\newblock In {\em Proceedings of the European Conference on Computer Vision
  (ECCV)}, pages 433--447, 2018.

\bibitem{zhu2019unsupervised}
Alex~Zihao Zhu, Liangzhe Yuan, Kenneth Chaney, and Kostas Daniilidis.
\newblock Unsupervised event-based learning of optical flow, depth, and
  egomotion.
\newblock In {\em Proceedings of the IEEE/CVF Conference on Computer Vision and
  Pattern Recognition}, pages 989--997, 2019.

\bibitem{ranccon2021stereospike}
Ulysse Ran{\c{c}}on, Javier Cuadrado-Anibarro, Benoit~R Cottereau, and
  Timoth{\'e}e Masquelier.
\newblock Stereospike: Depth learning with a spiking neural network.
\newblock {\em arXiv preprint arXiv:2109.13751}, 2021.

\bibitem{yu2020toward}
Zhaofei Yu, Jian~K Liu, Shanshan Jia, Yichen Zhang, Yajing Zheng, Yonghong
  Tian, and Tiejun Huang.
\newblock Toward the next generation of retinal neuroprosthesis: visual
  computation with spikes.
\newblock {\em Engineering}, 6(4):449--461, 2020.

\bibitem{9181055}
Jing Zhao, Ruiqin Xiong, and Tiejun Huang.
\newblock High-speed motion scene reconstruction for spike camera via motion
  aligned filtering.
\newblock In {\em 2020 IEEE International Symposium on Circuits and Systems
  (ISCAS)}, pages 1--5, 2020.

\bibitem{Zheng_2021_CVPR}
Yajing Zheng, Lingxiao Zheng, Zhaofei Yu, Boxin Shi, Yonghong Tian, and Tiejun
  Huang.
\newblock High-speed image reconstruction through short-term plasticity for
  spiking cameras.
\newblock In {\em Proceedings of the IEEE/CVF Conference on Computer Vision and
  Pattern Recognition (CVPR)}, pages 6358--6367, June 2021.

\bibitem{Zhao_2021_CVPR}
Jing Zhao, Ruiqin Xiong, Hangfan Liu, Jian Zhang, and Tiejun Huang.
\newblock Spk2imgnet: Learning to reconstruct dynamic scene from continuous
  spike stream.
\newblock In {\em Proceedings of the IEEE/CVF Conference on Computer Vision and
  Pattern Recognition (CVPR)}, pages 11996--12005, June 2021.

\bibitem{zheng2021high}
Yajing Zheng, Lingxiao Zheng, Zhaofei Yu, Boxin Shi, Yonghong Tian, and Tiejun
  Huang.
\newblock High-speed image reconstruction through short-term plasticity for
  spiking cameras.
\newblock In {\em Proceedings of the IEEE/CVF Conference on Computer Vision and
  Pattern Recognition}, pages 6358--6367, 2021.

\bibitem{zhu2021neuspike}
Lin Zhu, Jianing Li, Xiao Wang, Tiejun Huang, and Yonghong Tian.
\newblock Neuspike-net: High speed video reconstruction via bio-inspired
  neuromorphic cameras.
\newblock In {\em Proceedings of the IEEE/CVF International Conference on
  Computer Vision}, pages 2400--2409, 2021.

\bibitem{Zhao_2021_ICCV}
Jing Zhao, Jiyu Xie, Ruiqin Xiong, Jian Zhang, Zhaofei Yu, and Tiejun Huang.
\newblock Super resolve dynamic scene from continuous spike streams.
\newblock In {\em Proceedings of the IEEE/CVF International Conference on
  Computer Vision (ICCV)}, pages 2533--2542, October 2021.

\bibitem{howard2019searching}
Andrew Howard, Mark Sandler, Grace Chu, Liang-Chieh Chen, Bo~Chen, Mingxing
  Tan, Weijun Wang, Yukun Zhu, Ruoming Pang, Vijay Vasudevan, et~al.
\newblock Searching for mobilenetv3.
\newblock In {\em Proceedings of the IEEE/CVF international conference on
  computer vision}, pages 1314--1324, 2019.

\bibitem{Uhrig2017THREEDV}
Jonas Uhrig, Nick Schneider, Lukas Schneider, Uwe Franke, Thomas Brox, and
  Andreas Geiger.
\newblock Sparsity invariant cnns.
\newblock In {\em International Conference on 3D Vision (3DV)}, 2017.

\bibitem{Silberman:ECCV12}
Pushmeet~Kohli Nathan~Silberman, Derek~Hoiem and Rob Fergus.
\newblock Indoor segmentation and support inference from rgbd images.
\newblock In {\em ECCV}, 2012.

\bibitem{shen2021cfnet}
Zhelun Shen, Yuchao Dai, and Zhibo Rao.
\newblock Cfnet: Cascade and fused cost volume for robust stereo matching.
\newblock In {\em Proceedings of the IEEE/CVF Conference on Computer Vision and
  Pattern Recognition}, pages 13906--13915, 2021.

\bibitem{ronneberger2015u}
Olaf Ronneberger, Philipp Fischer, and Thomas Brox.
\newblock U-net: Convolutional networks for biomedical image segmentation.
\newblock In {\em International Conference on Medical image computing and
  computer-assisted intervention}, pages 234--241. Springer, 2015.

\bibitem{baheti2020eff}
Bhakti Baheti, Shubham Innani, Suhas Gajre, and Sanjay Talbar.
\newblock Eff-unet: A novel architecture for semantic segmentation in
  unstructured environment.
\newblock In {\em Proceedings of the IEEE/CVF Conference on Computer Vision and
  Pattern Recognition Workshops}, pages 358--359, 2020.

\bibitem{Chen_2021_CVPR}
Xiangyu Chen, Yihao Liu, Zhengwen Zhang, Yu~Qiao, and Chao Dong.
\newblock Hdrunet: Single image hdr reconstruction with denoising and
  dequantization.
\newblock In {\em Proceedings of the IEEE/CVF Conference on Computer Vision and
  Pattern Recognition (CVPR) Workshops}, pages 354--363, June 2021.

\bibitem{chen2021hinet}
Liangyu Chen, Xin Lu, Jie Zhang, Xiaojie Chu, and Chengpeng Chen.
\newblock Hinet: Half instance normalization network for image restoration.
\newblock In {\em Proceedings of the IEEE/CVF Conference on Computer Vision and
  Pattern Recognition}, pages 182--192, 2021.

\bibitem{Zhang_2019_CVPR}
Feihu Zhang, Victor Prisacariu, Ruigang Yang, and Philip~H.S. Torr.
\newblock Ga-net: Guided aggregation net for end-to-end stereo matching.
\newblock In {\em Proceedings of the IEEE/CVF Conference on Computer Vision and
  Pattern Recognition (CVPR)}, June 2019.

\bibitem{geiger2013vision}
Andreas Geiger, Philip Lenz, Christoph Stiller, and Raquel Urtasun.
\newblock Vision meets robotics: The kitti dataset.
\newblock {\em The International Journal of Robotics Research},
  32(11):1231--1237, 2013.

\bibitem{sim2021xvfi}
Hyeonjun Sim, Jihyong Oh, and Munchurl Kim.
\newblock Xvfi: Extreme video frame interpolation.
\newblock In {\em Proceedings of the IEEE/CVF International Conference on
  Computer Vision}, pages 14489--14498, 2021.

\end{thebibliography}

\end{document}